\documentclass[10pt,journal]{IEEEtran}
\usepackage{amsmath,amsfonts}
\usepackage{algorithmic}
\usepackage[ruled,linesnumbered]{algorithm2e}
\usepackage{array}
\usepackage{textcomp}
\usepackage{stfloats}
\usepackage{url}
\usepackage{verbatim}
\usepackage{graphicx}
\usepackage{cite}
\usepackage{boldline}
\usepackage{soul}
\usepackage{multirow}
\usepackage[hidelinks]{hyperref}
\usepackage{diagbox}
\usepackage{amssymb}
\usepackage{paralist,bm}
\usepackage{makecell}
\usepackage{subfigure}
\usepackage{xcolor}
\usepackage{booktabs}
\usepackage{times}
\usepackage{epsfig}
\usepackage{enumitem}
\usepackage{ragged2e}
\usepackage{float}
\usepackage{marvosym}

\graphicspath{{figs/}}

\def\ie{\emph{i.e.~}}

\def\etal{\emph{et al}.~}

\newenvironment{packed_itemize}{
	\begin{itemize}
		\setlength{\itemsep}{1pt}
		\setlength{\parskip}{0pt}
		\setlength{\parsep}{0pt}
	}{\end{itemize}}

\begin{document}
\title{Bridging Simulation-to-Reality Domain Gap via Topology-Aware Representation Learning}
\title{Topology-Aware Unsupervised Domain Adaptation for 3D Point Cloud Classification}
\title{Topology-Aware Modeling for Unsupervised Cross-Domain Point Cloud Recognition}
\title{Topology-Aware Modeling for Unsupervised Simulation-to-Reality Point Cloud Recognition}

\author{Longkun Zou${^*}$, Kangjun Liu${^*}$,
Ke Chen\textsuperscript{\Letter},~\IEEEmembership{Member,~IEEE}, 
Kailing Guo,~\IEEEmembership{Member,~IEEE},
Kui Jia,~\IEEEmembership{Member,~IEEE},
~and~Yaowei Wang,~\IEEEmembership{Member,~IEEE}
\IEEEcompsocitemizethanks{
\IEEEcompsocthanksitem This work is supported in part by the Guangdong Pearl River Talent Program (Introduction of Young Talent) under Grant No. 2019QN01X246, the Multi-source Cross-platform Video Analysis and Understanding forIntelligent Perception in Smart City under Grant No. U20B2052, the Guangdong Basic and Applied Basic Research Foundation under Grant No. 2023A1515011104. \textit{(Longkun Zou and Kangjun Liu contributed equally to this work.) (Corresponding author: Ke Chen.)}
\IEEEcompsocthanksitem L. Zou, K. Liu and K. Chen are with the Pengcheng Laboratory, Shenzhen 518000, China.
\IEEEcompsocthanksitem K. Guo is with the School of Electronic and Information Engineering, South China University of Technology, Guangzhou, 510641, China.
\IEEEcompsocthanksitem K. Jia is with the Chinese University of Hong Kong, Shenzhen (CUHK-Shenzhen), Shenzhen 518000, China.
\IEEEcompsocthanksitem Y. Wang is with the Harbin Institute of Technology, Shenzhen, China, and also with the Pengcheng Laboratory, Shenzhen, China.
}
}

\markboth{Journal of \LaTeX\ Class Files,~Vol.~XX, No.~X, MARCH~2024}
{Zou \MakeLowercase{\textit{et al.}}: Simulation-to-Reality Domain Adaptation on Point Clouds}

\IEEEpubid{0000--0000/00\$00.00~\copyright~2024 IEEE}

\IEEEtitleabstractindextext{
\begin{abstract}
\justifying{
Learning semantic representations from point sets of 3D object shapes is often challenged by significant geometric variations, primarily due to differences in data acquisition methods.
Typically, training data is generated using point simulators, while testing data is collected with distinct 3D sensors, leading to a simulation-to-reality (Sim2Real) domain gap that limits the generalization ability of point classifiers.
Current unsupervised domain adaptation (UDA) techniques struggle with this gap, as they often lack robust, domain-insensitive descriptors capable of capturing global topological information, resulting in overfitting to the limited semantic patterns of the source domain.
To address this issue, we introduce a novel Topology-Aware Modeling (TAM) framework for Sim2Real UDA on object point clouds.   
Our approach mitigates the domain gap by leveraging global spatial topology, characterized by low-level, high-frequency 3D structures, and by modeling the topological relations of local geometric features through a novel self-supervised learning task. 
Additionally, we propose an advanced self-training strategy that combines cross-domain contrastive learning with self-training, effectively reducing the impact of noisy pseudo-labels and enhancing the robustness of the adaptation process.
Experimental results on three public Sim2Real benchmarks validate the effectiveness of our TAM framework, showing consistent improvements over state-of-the-art methods across all evaluated tasks.
The source code of this work will be available at {\color{blue}\url{https://github.com/zou-longkun/TAG.git}}.
}
\end{abstract}
}
\maketitle
\IEEEdisplaynontitleabstractindextext

\begin{figure}[t]
  \centering
   \includegraphics[width=0.95\linewidth]{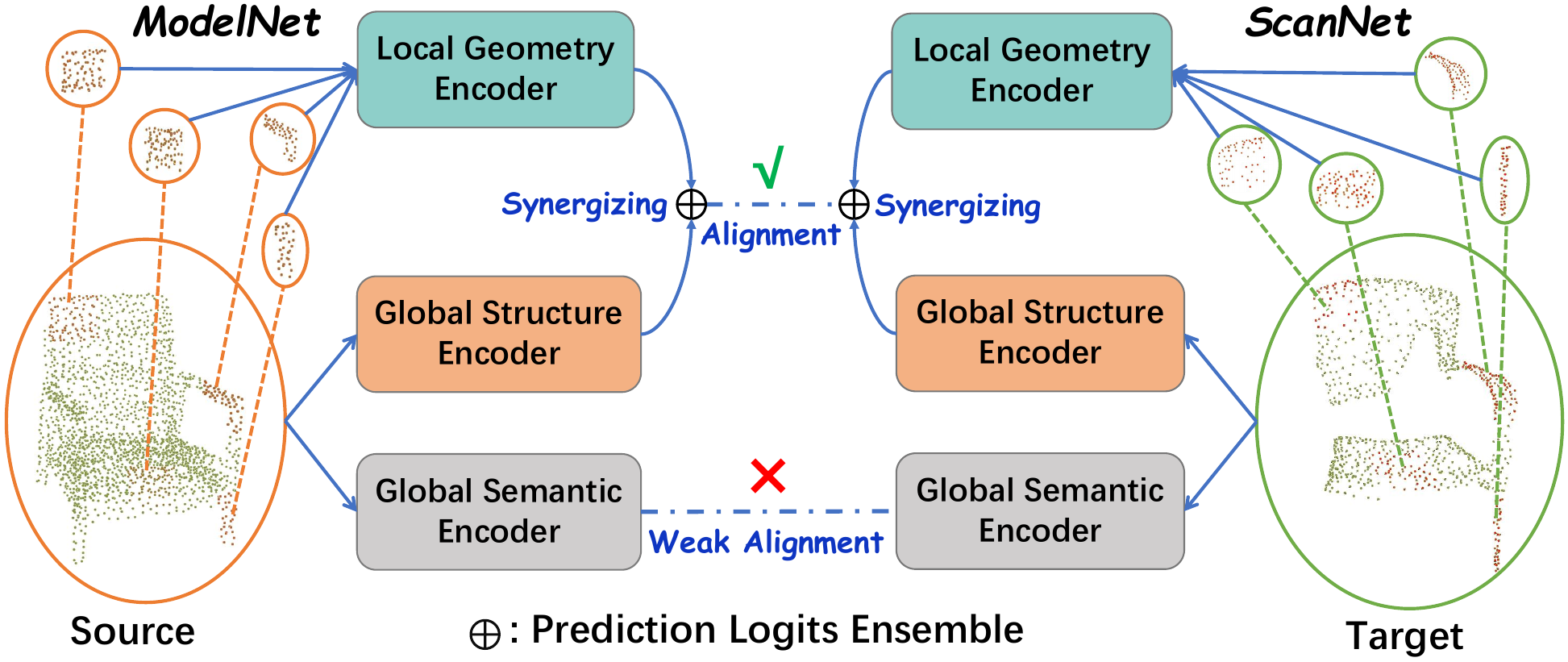}
   \caption{Illustration of alignment based on synergizing topological configuration of local geometric implicits and global spatial topology across domains. Aligning two point cloud objects from different domains may be difficult relying solely on high-level global semantics, but with knowledge of local geometric topological configuration and global spatial topology, alignment can be easily achieved.}
   \label{fig:part_align}
\end{figure}

\section{Introduction}

\IEEEPARstart{P}{} oint clouds are a widely utilized representation of 3D object shapes, valued for their simplicity in applications such as recognition \cite{PointGL, DAL_seg, CMNet, Geo_bp}, robotics \cite{Cattrack} and autonomous driving \cite{RTTracking, VPFNet, CenterTube}. The recent surge in semantic analysis of point clouds has been significantly influenced by synthetic datasets like ShapeNet \cite{Shapenet} and ModelNet \cite{Modelnet}, which have spurred the development of numerous algorithms for classification, segmentation, and detection \cite{Pointnet, Pointnet++, DGCNN, KPconv, PointASNLRP, PCT, Voxelnet, ModelingPC, SO_Net, SphericalC}. In contrast, research addressing real-world scenarios has focused on challenging datasets such as ScanNet \cite{Scannet} and ScanObjectNN \cite{ScanObjectNN}. Notably, Xu \etal \cite{PractClassification} have investigated the practical issues associated with single-view partial point clouds, utilizing adapted datasets like PartialScanNet and PartialScanObjectNN. These approaches generally operate under the assumption that both training and testing data adhere to the i.i.d. condition, an assumption that is often unrealistic in practical settings.

Meanwhile, the challenge of unsupervised domain adaptation (UDA) \cite{dan, can, dann, mcd, mstn, tpn} in the absence of an i.i.d. condition has garnered significant attention in recent years. The primary objective is to learn a shared representation space that facilitates the transfer of knowledge from a labeled source domain to an unlabeled, but related, target domain. Recent research has increasingly focused on UDA for point clouds, with notable works building upon the foundational PointDAN \cite{PointDAN} and extending into methods \cite{PCM_RegRecT, GAST, GAI, GLRV, QS3}. These point-based UDA approaches can generally be categorized into two main strategies: domain adversarial training \cite{PointDAN} and self-supervised learning \cite{GAST, PCM_RegRecT, GLRV, GAI}. Adversarial methods aim for explicit feature alignment across domains, adapting techniques from image-based UDA \cite{dann, mcd}, but they often suffer from instability and the risk of distorting the intrinsic data structures, leading to suboptimal adaptation. Conversely, self-supervised learning methods focus on designing pretext tasks to capture domain-invariant geometric patterns, such as rotation angle classification and deformation reconstruction \cite{GAST, PCM_RegRecT}, scaling prediction and 3D-2D-3D projection reconstruction \cite{GLRV}, and global implicit field learning \cite{GAI}. 
\IEEEpubidadjcol 
Although methods like PDG \cite{PDG} use part-level feature encoding to mitigate this, existing UDA algorithms for point clouds frequently struggle with overfitting to limited local semantic patterns, due to the absence of robust, domain-insensitive global topological descriptors. This challenge is compounded by the tendency of deep neural networks to prioritize local information over global context \cite{bakerdeep, geirhoshapebias}.

\begin{figure*}
  \centering
  \includegraphics[width=1.0\linewidth]{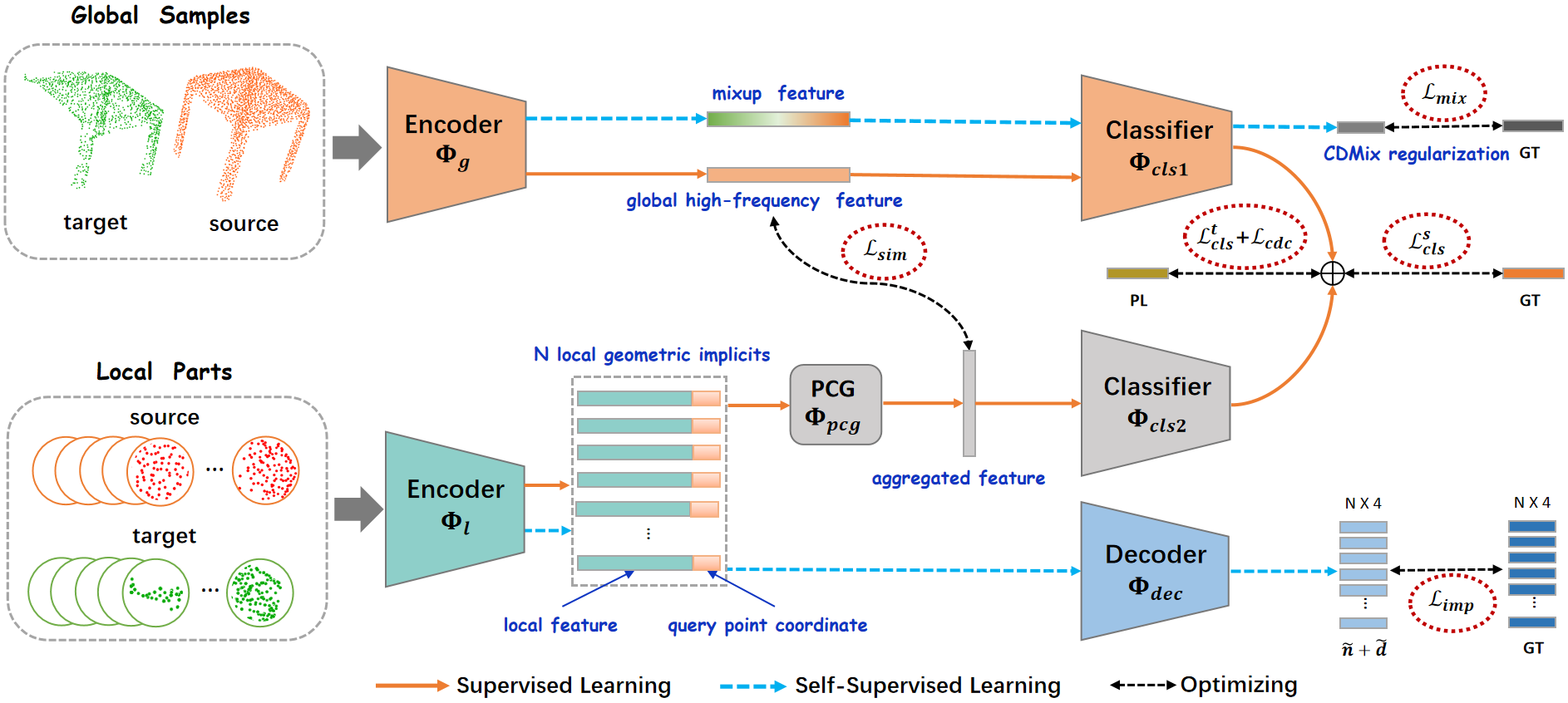}
  \caption{Overview of the proposed TAM framework, which consists of two branches: the top branch captures low-level global high-frequency 3D spatial structures, and the bottom branch learns domain-insensitive local geometric implicits through a self-supervised task. Note that the proposed framework has two types of input, global point clouds and their local parts. 
  Orange solid line indicates the supervised learning pipeline, which takes as input the global point cloud from the labeled source domain and is optimized with ground truth class labels. The blue dotted line represents the self-supervised learning pipeline, where the self-supervised Cross-Domain Mixup (CDMix) serves as a regularizer for the global feature branch, while local implicits branch takes as input the local part point clouds of the global point cloud from both source and target domains and is optimized with the surface projection signals (projection directions and distance to the approximate object surface in terms of the global point cloud) of query points. The Part-based Cloud Graph (PCG) module aggregates local geometric implicits into a global topological representation to exploit the relationships between local parts by supervision of ground truth class labels.}
  \label{fig:model}
\end{figure*}

In response to these challenges, cognitive neuroscience research \cite{Connecting, partsrecgnition} highlights the crucial role of global spatial topology and the interaction between local object parts in human visual recognition, a process notably different from how deep neural networks operate. Building on these insights, we argue that improving point cloud representations under distribution shifts requires a strong emphasis on both global spatial topology and the topological relationships of local geometric features. This leads to the development of a novel framework, \textbf{Topology-Aware Modeling (TAM)} as illustrated in Fig. \ref{fig:model}, tailored for Sim2Real UDA in object point clouds. Concretely, we first utilize Fourier Positional Encoding—comprising trigonometric functions—to capture low-level global high-frequency 3D spatial structures \cite{Point_NN, protein, nerf, fourier}, thereby revealing domain-insensitive global spatial topology. To further bolster the generalizability of this global representation, we propose a novel technique known as Cross-Domain Mixup (CDMix). This method introduces implicit regularization by creating convex combinations of training samples from different domains and enforcing consistency between predictions for these combined samples and the predictions of their original counterparts. Additionally, we assert that robust local geometric representations can uncover domain-agnostic semantic patterns related to the global topological configuration of shape primitives. Drawing inspiration from self-supervised learning approaches \cite{GAI}, we introduce a novel self-supervised learning framework for local geometric implicits, which are then synthesized into a global representation using a Part-based Cloud Graph (PCG), as illustrated in Fig. \ref{fig:part_align}.

Furthermore, building on recent advancements in \cite{GAST, GLRV, GAI}, we extend self-training with pseudo-labels to further address the domain gap. Given the inherent noisiness of pseudo-labels, their straightforward application often results in overfitting, especially to the dominant classes in the source domain. To counteract this, we propose a category-based cross-domain contrastive learning approach that emphasizes soft feature alignment rather than rigid final classification, thus mitigating the impact of noisy pseudo-labels and enhancing the robustness of self-training. 
Considering the challenges of manually annotating real-world 3D point clouds—typically acquired through depth cameras or LiDAR sensors—our approach is particularly relevant, as it provides a practical and effective solution to the Sim2Real adaptation problem, leveraging the substantial labeled synthetic point clouds available from CAD models \cite{Shapenet, Modelnet}.
We evaluate our method on three well-established point cloud benchmarks, namely PointDA-10 \cite{PointDAN}, Sim-to-Real \cite{MetaSets} and GraspNetPC-10 \cite{GAI}. The results consistently demonstrate the superiority of our proposed method over the current state-of-the-art in all Sim2Real tasks.

The primary contributions of this paper are summarized as follows:

\begin{packed_itemize}
    \item \textbf{A Novel Sim2Real UDA Framework:} We present a novel framework, termed TAM, for Sim2Real unsupervised domain adaptation on point cloud classification. The framework bridges the domain gap by learning domain-invariant topology-aware representations that capture global spatial topology and the topological relations of local geometric features.
    
    \item \textbf{Key Technical Innovations:} Our method utilizes Fourier Positional Encoding to capture global spatial topology, which is regularized via a proposed Cross-Domain Mixup (CDMix) strategy. Additionally, we introduce a self-supervised learning approach to model local geometric features, which are subsequently aggregated into a domain-agnostic global topological representation using a Part-based Cloud Graph (PCG) module.
    
    \item \textbf{Robust Self-Training:} We integrate cross-domain contrastive learning with self-training to mitigate the effects of noisy pseudo-labels, thereby improving the robustness of the learning process and minimizing overfitting to source-dominant classes.
    
    \item \textbf{State-of-the-Art Performance:} Extensive experiments on two public benchmarks demonstrate that our method consistently outperforms existing UDA approaches for point cloud classification in main Sim2Real tasks.
\end{packed_itemize}
\section{Related Works}

\noindent \textbf{Deep Classification on Point Clouds --}
To address the inherent sparsity and irregular distribution in point-based shape representations, numerous point cloud classification networks have been developed \cite{Pointnet,DGCNN,Pointnet++,tang2020improving,CMNet}. Many of these approaches rely on encoding features for individual points through multi-layer perceptrons (MLPs), followed by aggregating these features with max-pooling to generate a global, permutation-invariant shape descriptor. Other techniques, such as \cite{KPconv,PointCNN}, convert point clouds into structured representations like voxels and apply 3D convolution operations to extract meaningful features. Additionally, works like \cite{DGCNN,Clusternet} represent each point's local neighborhood as a spatial or spectral graph to encode local geometric relationships. Guo \etal \cite{PCT} leveraged transformers for semantic analysis on point clouds, setting new benchmarks in performance.
While these models achieve high accuracy in point cloud classification tasks, only a handful of works \cite{PointDAN,PCM_RegRecT,GAST,GAI,GLRV} have tackled the issue of closing the domain gap between different domains, particularly in Sim2Real unsupervised domain adaptation—a critical challenge this paper seeks to address.
 
\noindent \textbf{UDA on Point Cloud Classification --}
Despite extensive research on unsupervised domain adaptation (UDA) in 2D images \cite{dan,can,dann,mcd,mstn,tpn}, UDA for 3D point clouds remains challenging, particularly due to the difficulty in encoding domain-agnostic features from local geometries. Qin \etal \cite{PointDAN} were the first to explore UDA for point cloud classification, using domain adversarial training to align local and global features across domains. Achituve \etal \cite{PCM_RegRecT} introduced a self-supervised reconstruction approach that integrates domain-invariant local geometric patterns from partially distorted point clouds into a global representation, along with point cloud mixup to address imbalanced data. Zou \etal \cite{GAST} combined self-paced self-training, aimed at preserving target discrimination, with self-supervised learning for domain-invariant geometric feature extraction, capturing both global and local patterns consistent across domains. Fan \etal \cite{GLRV} proposed two auxiliary self-supervised tasks—scaling-up-down prediction and 3D-2D-3D projection reconstruction—and developed a reliable pseudo-label voting mechanism to further boost domain adaptation. Shen \etal \cite{GAI} applied a self-supervised task focused on learning geometry-aware global implicit representations for UDA on point clouds. In a different approach, Chen \etal \cite{QS3} combined realistic point cloud synthesis with quasi-balanced self-training, facilitating input and feature space alignment to mitigate domain gap issues.
Our method shares the underlying assumption from PDG in \cite{PDG} that simpler local geometries are more likely to be consistent across domains. However, unlike \cite{PDG}, we emphasize self-supervised learning of local geometric implicits, which are then aggregated into a topological global representation, rather than merely combining local features. This approach enhances the generalization ability for Sim2Real UDA in 3D point cloud tasks.

\noindent \textbf{Mixup-style Data Augmentation --}
Mixup-based data augmentation techniques have proven effective in training and regularizing neural networks, particularly in image-related tasks. For instance, Mixup \cite{mixup} creates interpolated samples by linearly blending two inputs, using the mixed data along with soft labels for training. Its successor, Manifold Mixup \cite{manifoldmixup}, extended this by performing interpolation in latent feature spaces to better exploit underlying visual information. CutMix \cite{cutmix} introduced the idea of inserting rectangular patches from one image into another to generate mixed data, while PuzzleMix \cite{puzzlemix} utilized saliency and local data statistics to propose a more sophisticated mixing strategy. Recently, Mixup has been explored for point cloud data. PointMixup \cite{PointMixup} applied the Mixup technique by computing one-to-one point correspondences between two point clouds, followed by interpolation of both the points and labels. PointAugment \cite{PointAugment} introduced an auto-augmentation framework that simultaneously optimized augmentation and classification networks through a complex adversarial process. PointCutMix \cite{PointCutmix} adapted CutMix for point cloud data. Despite these developments, the use of Mixup in unsupervised domain adaptation (UDA) for point cloud classification remains underexplored. In this work, we present a straightforward yet powerful cross-domain Mixup technique, serving as an implicit semantic regularization mechanism to promote more generalized representations in UDA for point cloud classification.


\section{Methodology}

In this paper, we focus on simulation-to-reality (Sim2Real) unsupervised domain adaptation on point cloud classification.
Consider a source domain \(\mathcal{S} = \{\mathcal{P}_i^s, y_i^s\}_{i=1}^{n_s}\) consisting of \(n_s\) labeled synthetic samples, and a target domain \(\mathcal{T} = \{\mathcal{P}_i^t\}_{i=1}^{n_t}\) containing \(n_t\) unlabeled real samples. Both domains share a common semantic label space \(\mathcal{Y}\) (i.e., \(\mathcal{Y}_{s} = \mathcal{Y}_{t}\)), where \(\mathcal{P} \in \mathcal{X} \subset \mathbb{R}^{N \times 3}\) denotes a point cloud comprised of \(N\) points with three-dimensional coordinates \((x,y,z)\). Our objective is to develop a domain-adapted mapping function \(\Phi: \mathcal{X} \rightarrow \mathcal{Y}\) that accurately classifies real target point cloud samples by leveraging the labeled synthetic data from the source domain and the unlabeled real data from the target domain. The mapping function $\Phi$ can be formulated into a cascade of a feature encoder $\Phi_\text{fea}: \mathcal{X} \rightarrow \mathbb{R}^d$ for any input $\mathcal{P}$ and a classifier $\Phi_\text{cls}: \mathbb{R}^{d}\rightarrow [0,1]^C$ typically using fully-connected layers as follows:
\begin{equation}
\begin{aligned}
&\Phi(\mathcal{P}) = \Phi_\text{cls}(\bm{z}) \circ \Phi_\text{fea}(\mathcal{P}), \\
&{\rm logits} = \Phi_\text{cls}(\bm{z}),
\end{aligned}
\end{equation}
where $d$ denotes the dimension of the feature representation output $\bm{z} \in \mathcal{Z} \subset \mathbb{R}^d$ of $\Phi_\text{fea}(\mathcal{P})$ and $C$ is the number of shared classes. 

In point-based Sim2Real UDA, source domain $\mathcal{S}$ consists of synthetic point clouds that are complete, uniform, and clean, while target domain $\mathcal{T}$ consists of real point clouds which are typically partial, non-uniform and noisy.
Therefore, there is a huge domain shift between $\mathcal{S}$ and $\mathcal{T}$ domains.

This paper proposes a novel scheme, termed Topology-Aware Modeling (TAM), for unsupervised domain adaptation on object point cloud classification, which bridges domain gap via learning domain-insensitive topology-aware representations from synthetic to real scenarios. We argue that the global spatial topology and the topological relations of local description are the keys to improving point representation under distribution shifts.
On the one hand, we utilize Fourier Positional Encoding composed of a set of trigonometric functions to capture low-level global high-frequency 3D spatial structures which intrinsically reveal domain-insensitive global spatial topology (see Sec. \ref{Point_PN}). Meanwhile, we design a novel self-supervised cross-domain mixup strategy as an implicit regularization for further favoring generalizable representations of global spatial topology that are robust to diverse topological variations across domains by maintaining consistency between predictions of mixed data and the combination of corresponding predictions of original samples (see Sec. \ref{CDPCM}).
On the other hand, we propose a novel self-supervised pretext task of learning local geometry-aware implicit fields that are shared across domains for encoding domain-agnostic local geometric patterns (see Sec. \ref{LGAIF}), which are then aggregated into a topological global representation in the form of a Part-based Cloud Graph (see Sec. \ref{PCG}). Furthermore, we propose to combine cross-domain contrastive learning and self-training to reduce the impact of noisy pseudo-labels and increase the robustness of self-training (see Sec. \ref{CLST}).


\subsection{Low-Level Global High-Frequency 3D Spatial Structures}
\label{Point_PN}
\subsubsection{Fourier Positional Encoding}
Existing point cloud data processing networks are usually designed as deep neural networks with a large number of trainable parameters, and the high-level semantic features they capture often overfit to limited local semantic patterns. Drawing from insights in cognitive neuroscience, we believe that the global spatial topology revealed by low-level spatial structure features is more generalizable across domains. In order to extract the domain-insensitive global high-frequency 3D spatial structure information of point cloud data, we utilize Fourier Positional Encoding—comprising trigonometric functions-following \cite{Point_NN, nerf, fourier}, denoted as $PosE(.)$, to transform the raw coordinates of points into high-dimensional vectors, and progressively aggregate local patterns via the multi-stage hierarchy. 
For a given point $\bm{p} = (x, y, z)$, the positional embedding function $PosE(.)$ maps it to a $d_0$-dimensional vector: 
\begin{eqnarray}
	\label{eq:trig_emb}
	\begin{aligned}
		PosE(p) = ConCat(\bm{f}^x, \bm{f}^y, \bm{f}^z) \in \mathbb{R}^{d_0},
	\end{aligned}
\end{eqnarray}
where $\bm{f}^x, \bm{f}^y, \bm{f}^z \in \mathbb{R}^{d_0/3}$ represent the embeddings along the $x$, $y$, and $z$ axes, respectively, and $d_0$ is the dimension of the initial feature vector. To illustrate, for the $x$-axis embedding $\bm{f}^x$, given a channel index $m \in [0, d_0/6]$, the embedding is computed as:
\begin{eqnarray}
	\label{eq:trig_emb_sin}
	\begin{aligned}
		\bm{f}^x[2m] = sine(\alpha x / \beta^{6m/d_0}),
	\end{aligned}
\end{eqnarray}
\begin{eqnarray}
	\label{eq:trig_emb_cos}
	\begin{aligned}
		\bm{f}^x[2m+1] = cosine(\alpha x / \beta^{6m/d_0}),
	\end{aligned}
\end{eqnarray}
where the parameters $\alpha$ and $\beta$ govern the scale and frequency, respectively. Leveraging the periodic properties of sine and cosine functions, this embedding effectively captures relative positional relationships between points and the detailed structural variations of 3D shapes. Following this embedding step, a four-stage network architecture, akin to PointNet++ \cite{Pointnet++}, is employed to hierarchically aggregate local spatial patterns.
For each stage includes: 

\textbf{(1) Extension}:
\begin{eqnarray}
	\label{eq:fea_exp}
	\begin{aligned}
		\bm{f}_{cj}=ConCat(\bm{f}_c, \bm{f}_j), j \in \mathcal{N}_c,
	\end{aligned}
\end{eqnarray}
where $\bm{f}_c$ represents the embedding of the point $\bm{p}_c$, which is sampled using Farthest Point Sampling (FPS), while $\mathcal{N}_c$ indicates the indices of the $k$ nearest neighbors of $\bm{p}_c$. Additionally, $\bm{f}_{cj}$ refers to the expanded feature corresponding to each neighboring point.

\textbf{(2) Interaction}:
\begin{eqnarray}
	\label{eq:geo_ext}
	\begin{aligned}
		\bm{f}^w_{cj}=(\bm{f}_{cj} + PosE(\Delta \bm{p}_j)) \odot PosE(\Delta \bm{p}_j), j \in \mathcal{N}_c,
	\end{aligned}
\end{eqnarray}
where $\Delta \bm{p}_j$ represents the normalized position of point $\bm{p}_j$, while $\bm{f}^w_{cj}$ refers to the weighted neighbor features, and $\odot$ indicates element-wise multiplication. The feature dimension of $PosE(\Delta \bm{p}_j)$ is aligned with that of $\bm{f}_{cj}$ across four stages, specifically $2d_0$, $4d_0$, $8d_0$, and $16d_0$, reflecting the progressive feature expansion.

\textbf{(3) Aggregation}:
\begin{eqnarray}
	\label{eq:fea_agg}
	\begin{aligned}
		\bm{f}^a_{cj}=MaxP(\{\bm{f}^w_{cj}\}_{j\in \mathcal{N}_c}) + AveP(\{\bm{f}^w_{cj}\}_{j\in \mathcal{N}_c}),
	\end{aligned}
\end{eqnarray}
where $MaxP$ and $AveP$ denote max pooling and average pooling operations, respectively. After processing through all four stages, a global representation $\bm{f}_g$ of the input point cloud is generated with a dimensionality of $d_1$. This global feature $\bm{f}_g$ is then projected into a $d$-dimensional space using a two-layer MLP projector, $\Phi_\text{proj}$. Finally, the projected features are passed through a classifier, $\Phi_\text{cls}$, to generate the final prediction logits.
\begin{eqnarray}
	\label{eq:logit_pn}
	\begin{aligned}
		&\Phi_\text{PNN}(\mathcal{P}) = \Phi_\text{cls}(\bm{z}_{g}) \circ \Phi_\text{proj}(\bm{f}_g) \circ \Phi_\text{g}(\mathcal{P}), \\
		&{\rm logits}_\text{PNN}=\Phi_\text{cls}(\bm{z}_{g}),
	\end{aligned}
\end{eqnarray}
where $\bm{z}_{g} \in \mathbb{R}^d$ of $\Phi_\text{proj}(\bm{f}_g)$ denotes the projected feature, $\Phi_\text{g}$ denotes the global feature encoder, $\Phi_\text{PNN}$ denotes the global high-frequency 3D spatial structures mapping function.

\subsubsection{Cross Domain Mixup on Point Clouds}
\label{CDPCM}
\begin{figure}
  \centering
  \includegraphics[width=0.95\linewidth]{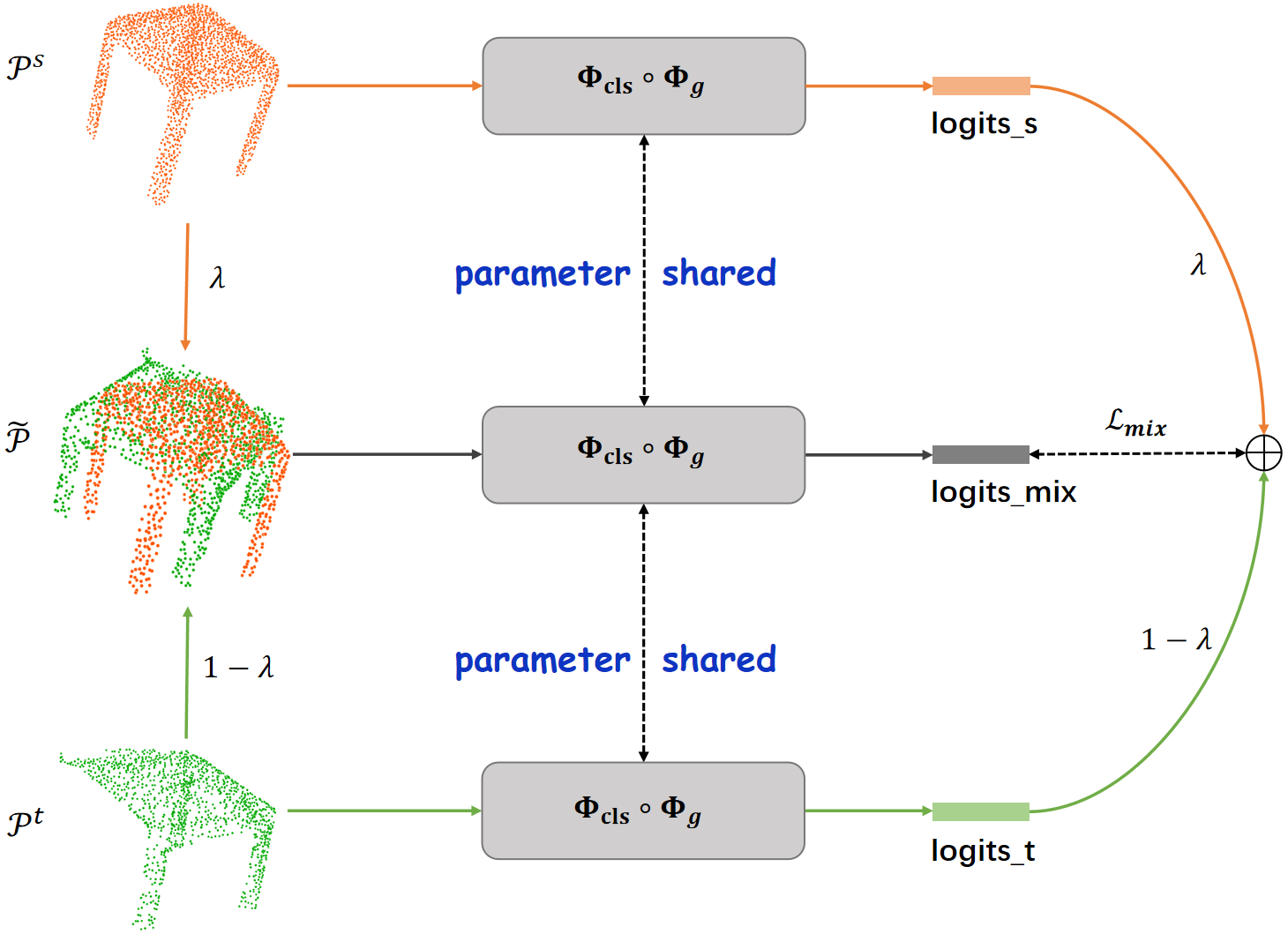}
  \vspace{-0.2cm}
  \caption{The cross-domain mixup (CDMix) strategy on point sets.}
  \label{fig:cdpcm}
\end{figure}
In general, Mixup enhances data by creating new labeled samples through convex combinations of pairs of data points and their respective labels. Several approaches have adapted Mixup for point clouds \cite{PointMixup, PCM_RegRecT}. Given two point cloud samples, $(\mathcal{P}_i, y_i)$ and $(\mathcal{P}_j, y_j)$, the process of point cloud Mixup is outlined as follows:
\begin{eqnarray}
\label{eq:mixup}
\begin{aligned}
&\widetilde{\mathcal{P}} = \lambda \mathcal{P}_i + (1 - \lambda)\mathcal{P}_j, \\
&\widetilde{y} = \lambda y_i + (1 - \lambda)y_j,
\end{aligned}
\end{eqnarray}
where $\lambda$ is the Mixup coefficient, which is randomly drawn from a beta distribution $Beta(\kappa, \kappa)$, where $\kappa \in (0, \infty)$. However, in unsupervised domain adaptation, the label information for the target domain is unavailable. Inspired by the approach in \cite{interconsist}, we replace the labels $y_i$ and $y_j$ with approximations $\Phi({\mathcal{P}_i})$ and $\Phi({\mathcal{P}_j})$, which represent the current predictions from the mapping function $\Phi = \Phi_\text{cls} \circ \Phi_\text{g}$. These predictions, referred to as virtual labels, are then used in our proposed cross-domain Mixup strategy as follows:
\begin{eqnarray}
\label{eq:cdpcm}
\begin{aligned}
&\widetilde{\mathcal{P}} = \lambda \mathcal{P}_i^s + (1 - \lambda)\mathcal{P}_j^t, \\
&\widetilde{y} = \lambda \Phi({\mathcal{P}_i^s}) + (1 - \lambda) \Phi({\mathcal{P}_j^t}),
\end{aligned}
\end{eqnarray}
where $\lambda \sim Beta(\kappa, \kappa)$, for $\kappa \in (0, \infty)$. We want the mapping function $\Phi$ to behave linearly along the lines between ${\mathcal{P}_i^s}$ and ${\mathcal{P}_j^t}$. Therefore, the prediction of $\widetilde{\mathcal{P}}$ should be consistent with $\widetilde{y}$. To this end, the following objective of CDMix can be given as:
\begin{eqnarray}
\label{eq:cpcm_loss}
\begin{aligned}
\min_{\Phi_{\text{g}}, \Phi_{\text{cls}}}\mathcal{L}_\text{mix} = CosSim(\Phi(\widetilde{\mathcal{P}}), \widetilde{y}).
\end{aligned}
\end{eqnarray}
Note that CDMix, designed for point cloud UDA, employs virtual labels similar to the method in \cite{interconsist}, yet it exhibits distinct characteristics. Specifically tailored for point cloud data, CDMix adopts a spatial-relation-based mixing strategy that bridges cross-domain scenarios, in contrast to the single-domain semi-supervised approach in \cite{interconsist}, which is designed for 2D image tasks. Moreover, CDMix can be applied to either the target or source domain, without utilizing the label information.
Although traditional conventional mixup strategies \cite{mixup, manifoldmixup, PointMixup, PointCutmix} are used for data augmentation, our proposed CDMix is designed without using label information, which is in the style of self-supervised learning. 

As an auxiliary implicit semantic regularization, we argue that CDMix can achieve stronger discriminability in the latent space while revealing more instances with diverse topological variations in the target domain. It helps the model enrich internal patterns in the shared feature space, leading to more continuous domain-invariant manifolds and thus helping to reduce the domain gap. 
Fig. \ref{fig:cdpcm} shows the procedure of our proposed CDMix.

\subsection{Neural Implicit Fields for Local Geometries}
\label{LGAIF}

\subsubsection{Query Points Sampling and Surface Projection}
\begin{figure}[t]
  \centering
   \includegraphics[width=0.95\linewidth]{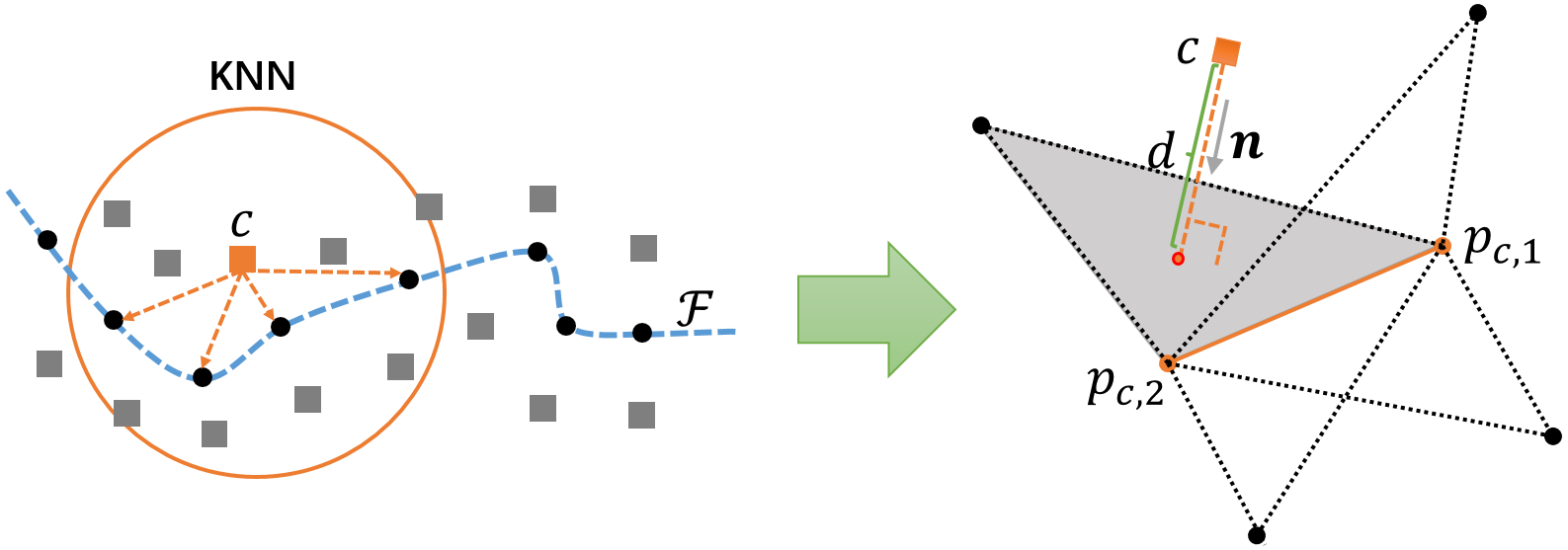}
   \caption{Illustration of query points sampling and surface projection. Black dots denote points in the point cloud, blue dashed lines denote the underlying surface $\mathcal{F}$ of the point cloud, and small squares represent query points distributed near the surface $\mathcal{F}$. Best viewed in color.}
   \label{fig:sss_signal}
\end{figure}

Existing neural networks tend to focus more on discriminative global semantic information and ignore local geometric information. Especially in the point-based UDA classification task, where there exists a huge domain discrepancy between the source synthetic and target real domains.
Shen \etal \cite{GAI} designed a self-supervised task of learning global geometry-aware implicits, which learns latent codes of 3D shapes by predicting the unsigned distance to the surface for each query point. 
An obvious observation is that the local geometric structures of point clouds are more likely to be shared across distinct domains, and thus being more generalizable to the geometry variations. 
Therefore, it is natural to make the network pay more attention to the extraction of local geometric information to improve cross-domain generalization.  Therefore, we design a self-supervised pretext task of learning local implicit fields for point-based Sim2Real UDA. 
As shown in Fig. \ref{fig:model}, the proposed module consists of two components, namely the Part-based Cloud Graph (PCG) module and the decoder $\Phi_\text{dec}$ respectively.

Given a point cloud, we divide the 3D space into equally spaced voxels. We define the resolution of a 3D voxel volume as $l^3$, the centers of voxels are equally distributed in the space, which serve as good candidates for query points, denoted as $\mathcal{C}_o$. Instead of using all centers, we choose the ones that are close to the underlying surface of the point cloud. The principle of choosing centers is that the distance between a center $\mathbf{c}$ and the surface $\mathcal{F}$ within a preset range: $Dist(\mathbf{c}, \mathcal{F}) \in [D_l, D_u]$. Since $\mathcal{F}$ is unknown, we approximate $Dist(\mathbf{c}, \mathcal{F})$ with a strategy: Firstly, find $M$ nearest points of $\mathbf{c}$ from the point cloud, denoted as $\{p_{c,1}, p_{c,2}, ..., p_{c,M}\}$ that are ordered from near to far. Next, from these points, we can form a set of triangles $\{T_m = (p_{c,1}, p_{c,2}, p_{c,m})\}_{m=3}^M$. Then, the $Dist(\mathbf{c}, \mathcal{F})$ can be approximated as follows:
\begin{equation}
Dist(\mathbf{c}, \mathcal{F}) \approx \min Dist(\mathbf{c}, \mathbf{t}), \mathbf{t} \in \{T_m\}_{m=3}^M,
\label{eq:query_point}
\end{equation}
where $\mathbf{t}$ represents a point contained in the constructed triangles. 

The next step is to search for the projection points on the surface of the query points. 
Note that when approximating $Dist(\mathbf{c}, \mathcal{F})$ using Equation (\ref{eq:query_point}), each query point yields an approximate nearest point $\mathbf{t_c}$ on the underlying surface of the point cloud. As a result, we take these points as projection points of the query points. Finally, we obtain the query point set $\mathcal{C} \in \mathcal{C}_o$. In our method, these projected points are utilized to construct self-supervised signals: 1) projection direction $\mathbf{n} \in [-1, 1]^3$ with $|\mathbf{n}|_1=1$, and 2) projection distance $d \in \mathbb{R}$, both defined from $\mathbf{c}$ to $\mathbf{t_c}$. Please refer to Fig. \ref{fig:sss_signal} for details.
To learn the local implicit fields, the decoder $\Phi_\text{dec}$ takes query point $\mathbf{c}$ and the feature of its $k$ nearest points in $\mathcal{P}$ as input:
\begin{equation}
ConCat(\widetilde{\mathbf{n}}, \widetilde{d}) = \Phi_\text{dec}(\bm{z}_c, \mathbf{c}) \circ \Phi_\text{l}(\mathcal{P}_c),
\end{equation}
where $\mathcal{P}_c = \{p_1,...,p_k\}$ denotes $k$ nearest points set of query point $\mathbf{c} \in \mathcal{C}$, and $\bm{z}_c \in \mathcal{Z}$ denotes the local feature representation output of $\Phi_\text{l}(\mathcal{P}_c)$. $ConCat(\widetilde{\mathbf{n}}, \widetilde{d})$ denotes concatenation of predicted 
projection direction and distance. 
The self-supervised loss for learning the local implicit field is as:
\begin{eqnarray}
\label{eq:imp_loss}
\begin{aligned}
\min_{\Phi_\text{l},\Phi_\text{dec}} \mathcal{L}_\text{imp} = - \frac{1}{N} \sum_{c=1}^{N} ||ConCat(\widetilde{\mathbf{n}}, \widetilde{d}) - ConCat(\mathbf{n}, d)||_2,
\end{aligned}
\end{eqnarray}
where $N$ is the number of sampled query points. 

\subsubsection{Aggregation on Local Implicit Representations}

\label{PCG}
\begin{figure}[t]
  \centering
   \includegraphics[width=0.95\linewidth]{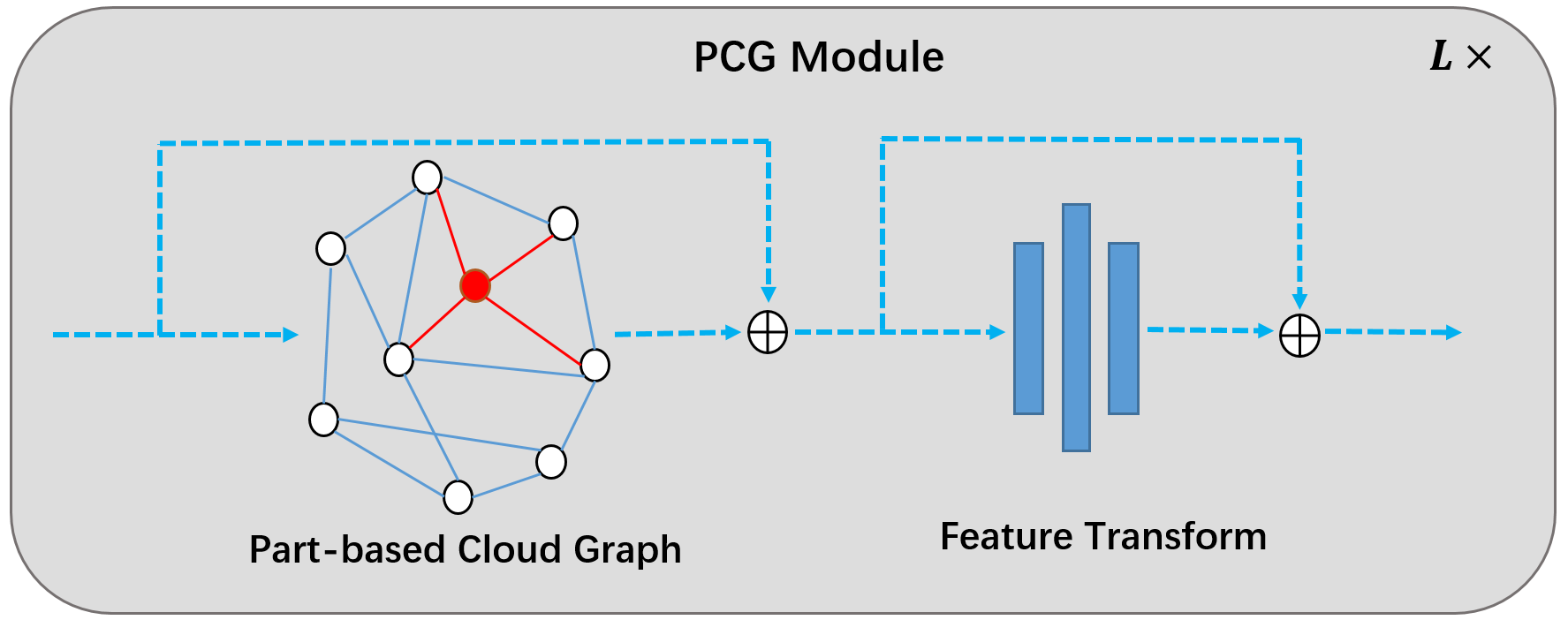}
   \caption{Illustration of graph construction of local implicit fields.}
   \label{fig:pcg}
\end{figure}

Vision GNN (ViGNN) \cite{ViG} reformulated the representation of 2D images as graph structures and leveraged graph-level features for various vision-related tasks. Inspired by the success of ViGNN, we introduce a Part-based Cloud Graph (PCG) module designed to capture relationships between different parts of the point cloud and integrate these part-level details into the global semantic representation. Unlike DGCNN \cite{DGCNN}, which models each individual point as a node, our approach segments the point cloud into multiple parts and treats each part as a distinct node. To achieve this, we sample $N$ query points (see Equation (\ref{eq:imp_loss})) and define each part by considering the $k$-nearest neighbors of each query point within the point cloud. We then apply $\Phi_\text{l}$ to encode each of these subsets into a feature vector $\bm{z}_i \in \mathbb{R}^d$, resulting in the feature set $\mathcal{Z} = {\bm{z}_1, \bm{z}_2, ..., \bm{z}_N}$. These features represent an unordered collection of nodes, expressed as $\mathcal{V} = {\bm{v}_1, \bm{v}_2, ..., \bm{v}_N}$. For each node $\bm{v}_i$, its $K$-nearest neighbors $\mathcal{N}(\bm{v}_i)$ are determined, and an edge $\mathbf{e}_{ji}$ is created, connecting $\bm{v}_j$ to $\bm{v}_i$ for all $\bm{v}_j \in \mathcal{N}(\bm{v}_i)$. The process of constructing this feature-based graph is denoted as $\mathcal{G} = G(\mathcal{Z})$. Once constructed, the module is used to propagate and exchange information across the nodes. As illustrated in Fig. \ref{fig:pcg}, the graph representation module includes two main components: a graph convolution network (GCN) module for processing graph-based information and a two-layer multi-layer perceptron (MLP) for node-level feature transformation. The graph convolution specifically operates as follows:
\begin{eqnarray}
\label{eq:gcnn_update}
\begin{aligned}
\mathcal{Z}^{'} = Update(Agg(\mathcal{G}, W_{agg}), W_{update}),
\end{aligned}
\end{eqnarray}
where $W_{agg}$ and $W_{update}$ are learnable weights of the aggregation and update operations, respectively. More specifically, the aggregation step generates a node's representation by integrating the features from its neighboring nodes:
\begin{eqnarray}
\label{eq:aggration}
\begin{aligned}
\bm{z}_{i}^{'} = h(\bm{z}_i, g(\bm{z}_i,\mathcal{N}(\bm{z}_i), W_{agg}), W_{update}),
\end{aligned}
\end{eqnarray}
where $\mathcal{N}(\bm{z}_{i})$ represent the set of neighboring nodes for $\bm{z}_{i}$. In this context, we utilize max-relative graph convolution \cite{deepgcns} due to its simplicity and efficiency:
\begin{eqnarray}
\label{eq:max_relative_gcnn}
\begin{aligned}
& \bm{z}_{i}^{''} = g(.) = [\bm{z}_i, \max(\{ \bm{z}_j W_{agg} - \bm{z}_i|j \in \mathcal{N}(\bm{z}_i)\})], \\
& \bm{z}_{i}^{'} = h(.) = \bm{z}_{i}^{''} W_{update},
\end{aligned}
\end{eqnarray}
where the bias term is omitted, the graph-level processing mentioned above can be expressed as: $\{\bm{z}_{i}^{'}\}_{i=1}^N =\mathcal{Z}^{'} = GraphConv(\mathcal{Z})$. Following ViGNN \cite{ViG}, we also incorporate a linear layer before and after graph convolution to project node features into the same domain and increase the feature diversity as follows:
\begin{eqnarray}
\label{eq:aggration_linear}
\begin{aligned}
F = \sigma (GraphConv(\mathcal{Z}W_{1}))W_{2} + \mathcal{Z},
\end{aligned}
\end{eqnarray}
where $F \in \mathbb{R}^{N \times d}$, $W_{1}$ and $W_{2}$ are weights of two fully-connected layers, $\sigma$ is the nonlinear activation function, and the bias term is omitted. To further encourage the feature transformation capacity and relieve the over-smoothing phenomenon, we additionally add a MLP with two fully-connected layers on each node:
\begin{eqnarray}
\label{eq:aggration_maxpool}
\begin{aligned}
\bm{z}_{pcg} = MaxP(\sigma (FW_{3})W_{4} + F),
\end{aligned}
\end{eqnarray}
where $\bm{z}_{pcg} \in \mathbb{R}^{d}$ denotes the aggregated feature that integrates all node information, $W_{3}$ and $W_{4}$ are weights of fully-connected layers, and the bias term is omitted, $MaxP$ denotes max pooling operation. We denote the encoding process of the above graph structure representation module as $\bm{z}_{pcg} = \Phi_{\text{pcg}}(\mathcal{P})$. Then we use the cosine similarity loss to further regularize feature $\bm{z}_{pcg}$ to align with global feature $\bm{z}_{g}$ of $\Phi_\text{g}(\mathcal{P})$:
\begin{eqnarray}
\label{eq:pcg_sim_loss}
\begin{aligned}
\min_{\Phi_{\text{l}}, \Phi_{\text{pcg}}}\mathcal{L}_\text{sim} = CosSim(\bm{z}_{pcg}, \bm{z}_{g}),
\end{aligned}
\end{eqnarray}
where $CosSim$ is the cosine similarity function. 
We finally feed $\bm{z}_{pcg}$ to a classifier $\Phi_\text{cls}$ to obtain the prediction logits of the aggregated structured graph representation.
\begin{eqnarray}
\label{eq:logit_pcg}
\begin{aligned}
&\Phi_\text{PCG}(\mathcal{P}) = \Phi_\text{cls}(\bm{z}_{pcg}) \circ \Phi_\text{pcg}(\mathcal{P}), \\
&{\rm logits}_\text{PCG}=\Phi_\text{cls}(\bm{z}_{pcg}),
\end{aligned}
\end{eqnarray}
where $\Phi_{\text{PCG}}$ denotes the mapping function for part-based cloud graph structure.

After regularization by PCG and CDMIX modules, a regularized high-level global semantic feature $\bm {z}_{reg} \in \mathbb {R}^d$ is obtained by $\Phi_\text{reg}(\bm {z}_{g})$. This process can be expressed as follows:
\begin{eqnarray}
\label{eq:fea_reg}
\begin{aligned}
&\Phi_\text{REG} = \Phi_\text{cls}(\bm {z}_{reg}) \circ \Phi_\text{reg}(\bm {z}_{g}) \circ \Phi_\text{proj}(\bm{f}_g) \circ \Phi_\text{g}(\mathcal{P}),\\
&{\rm logits}_\text{REG}=\Phi_\text{cls}(\bm{z}_{reg}),
\end{aligned}
\end{eqnarray}
where $\Phi_\text{REG}$ denotes regularized the global high-level semantic mapping function, \(\Phi_\text{reg}\) denotes the global feature encoder \(\Phi_\text{g}\), regularized by CDMix and PCG moudules.

\subsection{Self-Train with Cross Domain Contrastive Learning}
\label{CLST}
Denote the labeled source samples $\{\mathcal{P}_i^s, y_i^s\}_{i=1}^{n_s}$ and their category probability vectors tuple $\{({\bm {{\rm p}}_1}_i^s, {\bm {{\rm p}}_2}_i^s)\}_{i=1}^{n_s}$ predicted by two different models $\Phi_\text{REG}, \Phi_\text{PCG}$ respectively, which are trained via minimizing the cross-entropy loss: 
\begin{eqnarray}
\label{EqnSrcSupLearn}
\begin{aligned}
\min_{\Phi_\ast} \mathcal{L}_\text{cls}^s = - \frac{1}{n_s} \sum_{i=1}^{n_s} \sum_{c=1}^{C} {\rm I}[c = y_i^s] \log {\rm p}_{i,c}^s,
\end{aligned}
\end{eqnarray}
where $\Phi_\ast = \{\Phi_\text{REG}, \Phi_\text{PCG}\}$, and ${\rm p}_{i,c}^s = {{\rm p}_1}_{i,c}^s{{\rm p}_2}_{i,c}^s$, with ${{\rm p}_1}_{i,c}^s$ and ${{\rm p}_2}_{i,c}^s$ representing the $c$-th elements of the category predictions ${\bm {{\rm p}}_1}_i^s$ and ${\bm {{\rm p}}_2}_i^s$ for a source point cloud $\mathcal{P}_i^s$, respectively. ${\rm I[\cdot]}$ is an indicator function. 
Supervised learning establishes in feature space $\mathcal{Z}$ two semantic representations $\bm{z}_{reg}, \bm{z}_{pcg}$ that are discriminative among categories on source domain $\mathcal{S}$.

\begin{algorithm}[htb]
\SetKwInOut{Input}{Input}
\SetKwInOut{Output}{Output}
\SetKwInOut{Parameter}{Parameter}
\SetKwInOut{Initialization}{Initialization}
\Input{labeled source data $\mathcal{S} = \{\mathcal{P}_i^s, y_i^s\}_{i=1}^{n_s}$, \\
       unlabeled target data $\mathcal{T} = \{\mathcal{P}_i^s\}_{i=1}^{n_t}$, \\
       local geometry mapping function $\Phi_\text{PCG}$,\\
       \mbox {global semantic mapping function $\Phi_\text{REG}$,}\\
       \mbox {source global feature bank $\mathcal{B} = \{\bm{z}_i^s\}_{i=1}^{n_s}$,}\\
       number of training rounds $R$, \\
       number of epochs $E$ for each round.\\
       }
  \Output{$\Phi_\text{PCG}$ and $\Phi_\text{REG}$}
  \Initialization {pretrained $\Phi_\text{PCG}$, $\Phi_\text{REG}$;\\
                   $\theta_0 = 0.8.$}
  \For{$r\leftarrow 1$ \KwTo $R$}{
    \For{$e\leftarrow 1$ \KwTo $E$}{
        \For{$(\mathcal{P}_i^s, y_i^s), (\mathcal{P}_i^t)$ in $(\mathcal{S}, \mathcal{T})$}{
            $\min_{\Phi_\text{PCG}, \Phi_\text{REG}} \mathcal{L}_\text{cls}^s$ with $(\mathcal{P}_i^s, y_i^s)$;\\
            $\min_{\Phi_\text{REG}} \mathcal{L}_\text{mix}$ with $(\mathcal{P}_i^s, y_i^s)$ and $\mathcal{P}_i^t$;\\
            $\min_{\Phi_\text{REG}} \mathcal{L}_\text{imp}$ with $(\mathcal{P}_i^s, y_i^s)$ and $\mathcal{P}_i^t$;\\
            $\min_{\Phi_\text{REG}} \mathcal{L}_\text{sim}$ with $(\mathcal{P}_i^s, y_i^s)$ and $\mathcal{P}_i^t$;\\
        }
        \For{$\mathcal{P}_i^t$ in $\mathcal{T}$}{
            \eIf {$h_i = 1$}{
                $\min_{\Phi_\text{PCG}, \Phi_\text{REG}} \mathcal{L}_\text{cls}^t$ with $(\mathcal{P}_i^t, \widehat{\bm{y}}_i^t)$;\\
                $\min_{\Phi_\text{REG}} \mathcal{L}_\text{cdc}$ with $(\mathcal{P}_i^t, \widehat{\bm{y}}_i^t)$ and $\mathcal{B}$;\\
            }{
            continue;
            }
       }
    } 
    $\theta_r = \theta_{r-1} + \epsilon$; \\
    update pseudo labels $\{\widehat{\bm{y}}_i^t\}_{i=1}^{n_t}$ and selection indicator $\bm{h}$ based on $\Phi_\text{PCG}$, $\Phi_\text{REG}$ and $\theta_r$.
  }
  \caption{Self-Training with Contrastive Learning}
  \label{clst}
\end{algorithm}

Self-paced self-training (SPST) is a widely used approach for aligning two domains by producing pseudo-labels in the target domain based on high-confidence predictions. Building on previous studies \cite{GAST, GLRV, GAI}, we adopt the SPST strategy to further minimize domain shift. The objective of self-paced learning-based self-training can be formulated as follows:
\begin{align}
\label{EqnTrgSelfLearn}
\begin{aligned}
\min_{\Phi_\ast, \widehat{\bm{Y}}^t} \mathcal{L}_\text{cls}^t = - \frac{1}{\widehat{n}_t} \sum_{i=1}^{\widehat{n}_t} \left(  \sum_{c=1}^C \widehat{y}_{i,c}^t \log {\rm p}_{i,c}^t + \gamma |\widehat{\bm{y}}_i^t|_1 \right),
\end{aligned}
\end{align}
where $\widehat{\bm{y}}_i^t$ represents the predicted pseudo-label as a one-hot vector for a target instance $\mathcal{P}_i^t$, where $\widehat{y}_{i,c}^t$ denotes its $c$-th component. The set of all pseudo-label vectors is denoted as $\widehat{\bm{Y}}^t = \{\widehat{\bm{y}}_i^t\}_{i=1}^{\widehat{n}_t}$. Additionally, ${\rm p}_{i,c}^t = {{\rm p}_1}_{i,c}^t{{\rm p}_2}_{i,c}^t$ follows a structure similar to ${\rm p}_{i,c}^s$, and $\gamma$ serves as a hyper-parameter that governs the number of target samples selected—the larger the value of $\gamma$, the more samples are chosen. The parameter $\gamma$ can be directly converted into a prediction confidence threshold $\theta_0 = \exp(-\gamma)$. When the parameters of $\Phi_\ast$ are fixed, the general pseudo-labeling strategy can be simplified into the following form:
\begin{eqnarray}
\label{LabelAssign}
\widehat{y}^t_{i,c} =\! \left\{
    \begin{aligned}
    & 1, \:\: {\rm if} \: c = \arg\max_{c} p(c|{\rm logits}) \: \text{and} \ p(c|{\rm logits}) > \theta_0, \\
    & 0, \:\: {\rm otherwise},
    \end{aligned}
    \right.
\end{eqnarray}
where ${\rm logits} = Avg({\rm logits}_\text{REG}, {\rm logits}_\text{PCG})$. The selection indicator $\bm{h}$ can be simply defined as $h_i = |\widehat{\bm{y}}_i^t|_1$. We set a threshold $\theta_0$ that grows gradually with each self-paced round, increasing by a constant $\epsilon$.

However, this type of supervision strongly directs the mapping function $\Phi$ to place $\{\mathcal{P}_i^t\}_{i=1}^{\widehat{n}_t}$ into the space of $\widehat{\bm{Y}}^t$, making it vulnerable to noisy labels. Inspired by the method in \cite{CDCDA}, we incorporate category-based cross-domain contrastive learning for pseudo-labeled samples during the self-training phase. This approach is akin to the soft nearest neighbor loss concept \cite{nonl_embed, impgen}, which encourages soft feature alignment instead of enforcing strict classification. Consequently, it offers greater resilience against potential errors in pseudo-labels. Recent research further confirms that supervised contrastive losses \cite{SCL} demonstrate improved robustness compared to traditional cross-entropy losses in image classification tasks. In our case, we leverage this property and demonstrate that cross-domain contrastive learning can mitigate the effects of noisy pseudo-labels, thus enhancing the robustness of self-training. The objective of this category-based cross-domain contrastive learning can be formulated as:

\begin{align}
\label{EqnCDContra}
\begin{aligned}
\min_{\Phi_\text{REG},\widehat{\bm{Y}}^t} \mathcal{L}_\text{cdc} = - \log \frac{\sum_{i=1}^{\widehat{n}_t} \phi(\bm{z}_{i}^t, \bm{z}^{s+})}{\sum_{i=1}^{\widehat{n}_t} \phi(\bm{z}_{i}^t, \bm{z}^{s+}) + \phi(\bm{z}_{i}^t, \bm{z}^{s-})}, 
\end{aligned}
\end{align}
where $\phi(.)$ denotes the similarity measurement function (\ie cosine similarity function), $\bm {z}^{s+}$ are the high-level global semantic features of positive samples of source domain with the same category label as $\bm{z}_i^t$,  while $\bm {z}^{s-}$ are the high-level global semantic features of negative samples with different category labels, $\mathcal{B} = \{\Phi_\text{REG}(\mathcal{P}_i^s)\}_{i=1}^{n_s} = \{\bm {z}^{s+}, \bm {z}^{s-}\}$. For convenience, we refer to the self-training combined with cross-domain contrastive learning as CLST.  More details are given in Algorithm \ref{clst}.

\subsection{Overall Loss}
The overall training objective integrates both semantic representation learning and geometry-aware regularization, leading to a unified framework for Sim2Real UDA on point clouds, as follows:
\begin{eqnarray}\label{EqnOverallObj}
\begin{aligned}
\min_{\Phi_\ast, \widehat{\bm{Y}}^t,\Phi_\text{dec}} \mathcal{L} =\mathcal{L}_\text{cls}^s + \lambda_t \mathcal{L}_\text{cls}^t + \lambda_{cdc} \mathcal{L}_\text{cdc} + \\
\lambda_{imp} \mathcal{L}_\text{imp} + \lambda_{mix} \mathcal{L}_\text{mix} + \lambda_{sim} \mathcal{L}_\text{sim},
\end{aligned}
\end{eqnarray}
where $\lambda_t, \lambda_{cdc}, \lambda_{imp}, \lambda_{mix}$ and $\lambda_{sim}$ are trade-off hyper-parameters, empirically set to 1, 1, 1, 1 and 0.1 respectively, and all model parameters of the model are simultaneously learned in an end-to-end manner. Once trained, our model can be deployed as a standard classification model by discarding all self-supervised task pipelines (\ie, the blue dotted lines in Fig. \ref{fig:model}) and retaining only the supervised task pipelines (\ie, the orange solid lines in Fig. \ref{fig:model}).

\section{Experiment}
\subsection{Datasets}
\noindent \textbf{PointDA-10.} PointDA-10 \cite{PointDAN} is a commonly used benchmark for domain adaptation on point clouds, featuring object point clouds from 10 shared categories across three datasets: ModelNet40 \cite{Modelnet}, ShapeNet \cite{Shapenet}, and ScanNet \cite{Scannet}. The ModelNet-10 (\textbf{M10}) dataset consists of 4,183 training and 856 testing point clouds, while ShapeNet-10 (\textbf{S10}) contains 17,378 training and 2,492 testing point clouds. Both datasets are constructed by sampling points from the surfaces of clean CAD models as described in \cite{Pointnet++}, with ModelNet-10 using 2048 points per sample and ShapeNet initially using 1024 points. To maintain consistency, we re-sampled ShapeNet-10 to also contain 2048 points per sample. Notably, the original ShapeNet-10 omitted a significant number of plant class samples and a smaller number of table class samples, but our version fully retains all samples, increasing the complexity of the dataset. The re-sampled ShapeNet-10 (\textbf{S10}) now comprises 17,627 training and 2,549 testing point clouds.
ShapeNet-10 is inherently more diverse compared to ModelNet-10 due to the larger variety of object instances and their structural differences. In contrast, ScanNet-10 (\textbf{S*10}) is made up of real-world object point clouds derived from scanned and reconstructed scenes, which are often noisy and incomplete due to occlusions and sensor imperfections. ScanNet-10 consists of 6,110 training and 1,769 testing samples, each containing 2,048 points collected from partially visible objects within manually annotated bounding boxes. We adhered to the data preparation procedure outlined in \cite{PCM_RegRecT, GAST}, aligning point clouds across all domains (datasets) along the gravity axis, allowing arbitrary rotations along the $z$-axis. Furthermore, each sample was downsampled to 1,024 points using farthest point sampling and normalized to a unit scale. With these datasets, our focus is on Sim2Real unsupervised domain adaptation in two tasks: \textbf{M10} $\rightarrow$ \textbf{S*10} and \textbf{S10} $\rightarrow$ \textbf{S*10}.

\noindent \textbf{Sim-to-Real.} The Sim-to-Real dataset \cite{MetaSets} serves as a relatively new benchmark designed to address the challenges of 3D domain generalization (3DDG). It comprises object point clouds from 11 shared categories between ModelNet40 \cite{Modelnet} and ScanObjectNN \cite{ScanObjectNN}, as well as 9 shared categories between ShapeNet \cite{Shapenet} and ScanObjectNN. This dataset is structured into four subsets: ModelNet-11 (\textbf{M11}), ScanObjectNN-11 (\textbf{SO*11}), ShapeNet-9 (\textbf{S9}), and ScanObjectNN-9 (\textbf{SO*9}).
Specifically, \textbf{M11} contains 4,844 training samples and 972 testing samples, while \textbf{SO*11} includes 1,915 training samples and 475 testing samples. In the ShapeNet-9 subset, \textbf{S9} comprises 19,904 training samples and 1,995 testing samples, whereas \textbf{SO*9} offers 1,602 training samples and 400 testing samples.
Following the protocol in \cite{GLRV}, two Sim2Real domain adaptation scenarios are commonly explored: \textbf{M11} $\rightarrow$ \textbf{SO*11} and \textbf{S9} $\rightarrow$ \textbf{SO*9}. These tasks evaluate cross-domain generalization by adapting models from clean synthetic datasets (ModelNet and ShapeNet) to noisy, incomplete real-world data (ScanObjectNN). This benchmark effectively tests the robustness of domain generalization methods in 3D object recognition.

\noindent \textbf{GraspNetPC-10.} The GraspNetPC-10 dataset \cite{graspnet} is an extension of GraspNet, which is generated by projecting raw depth scans from Kinect2 and Intel Realsense cameras into 3D space and cropping them with segmentation masks. It includes both synthetic and real-world point clouds across 10 classes. Specifically, the synthetic domain (\textbf{Syn.}) contains 12,000 training point clouds, while the Kinect domain (\textbf{Kin.}) offers 10,973 training point clouds and 2,560 testing point clouds. The Realsense domain (\textbf{RS.}) consists of 10,698 training point clouds and 2,560 testing point clouds. The real-world data from these two devices are influenced by various noise factors, geometric distortions, and missing parts. 
Four domain adaptation scenarios are commonly explored: \textbf{Syn.}$\rightarrow$\textbf{Kin.}, \textbf{Syn.}$\rightarrow$\textbf{RS.}, \textbf{Kin.}$\rightarrow$\textbf{RS.} and \textbf{RS.}$\rightarrow$\textbf{Kin.}.

\subsection{Implementation Details}
We adopt the Point-PN \cite{Point_NN} as the backbone of the \textit{Global Feature Encoder} $\Phi_\text{g}$ and the DGCNN \cite{DGCNN} as backbone of the \textit{Local Feature Encoder} $\Phi_\text{l}$, while the \textit{Category Classifier} $\Phi_\text{cls}$ is based on a multi-layer perceptron (MLP) with three fully-connected layers. \textit{Decoder} $\Phi_\text{dec}$ is a four-layers MLP (\ie $\{2051, 512, 128, 4\}$) followed by ReLU activation function.
By default, the hyper-parameters of $\alpha, \beta, \kappa, \lambda_t, \lambda_{cdc}, \lambda_{imp}, \lambda_{mix} $ and $\lambda_{sim}$ are empirically set to 100, 500, 2, 1, 1, 1, 1 and 0.1 respectively. For the training process, we use the Adam optimizer \cite{adam} with a constant learning rate of 0.0001 for learning local implicit geometries in a self-supervised manner. For both local and global supervised semantic learning, an initial learning rate of 0.001 is applied, along with a cosine annealing scheduler that adjusts the learning rate on a per-epoch basis. We incorporate dropout at a rate of 0.5 and batch normalization \cite{batchnorm} after both the convolutional and hidden layers as needed. The batch size is set to 16 for 200 training epochs. During self-training, a cosine annealing scheduler adjusts the learning rate over 10 epochs for each round. By default, the self-training rounds $R$ and $\epsilon$ are set to 5 and 0.005, respectively.

\begin{table}[htbp]
  \caption{Classification accuracy (\%) averaged over 3 seeds ($\pm$ SEM) on the PointDA-10 dataset. M10: ModelNet-10; S10: ShapeNet-10; S*10: ScanNet-10. We concentrate on Sim2Real scenarios: M$\rightarrow$S* and S$\rightarrow$S* and our method achieves best performance. {$\dagger$} denotes experiments without using 3 seeds.}
  \label{tab:pointda}
  \centering
  \resizebox{0.9\linewidth}{!}{
      \begin{tabular}{lc||cc}
      \toprule
      Method & SPST & M10$\rightarrow$S*10 & S10$\rightarrow$S*10 \\
      \midrule
      w/o Adaptation && 43.8$\pm$2.3 & 42.5$\pm$1.4 \\
      \midrule
      DANN \cite{dann} && 42.1$\pm$0.6 & 50.9$\pm$1.0 \\
      PointDAN \cite{PointDAN} && 44.8$\pm$1.4 & 45.7$\pm$0.7 \\
      RS \cite{SSL_PointSeg_PC} && 46.7$\pm$4.8 & 51.4$\pm$3.9 \\
      DefRec \cite{PCM_RegRecT} && 46.6$\pm$2.0 & 49.9$\pm$1.8 \\
      DefRec+PCM \cite{PCM_RegRecT} && 51.8$\pm$0.3 & 54.5$\pm$0.3 \\
      \multirow{2}{*}{GAST \cite{GAST}}
      && 56.7$\pm$0.3 & 55.0$\pm$0.2\\
      &\checkmark& 59.8$\pm$0.1 & 56.7$\pm$0.2\\
      \multirow{2}{*}{GAI \cite{GAI} }
      && 55.3$\pm$0.3 & 55.4$\pm$0.5\\
      &\checkmark& 58.6$\pm$0.1 & 56.9$\pm$0.2\\
      GLRV \cite{GLRV} &\checkmark& 60.4$\pm$0.4 & 57.7$\pm$0.4 \\
      PDG \cite{PDG} && 57.9$\pm$0.3 & 50.0$\pm$0.2 \\
      SD{$^\dagger$} \cite{SD} &\checkmark& 61.1$\pm$0.0 & 58.9$\pm$0.0 \\
      \multirow{2}{*}{DAS \cite{DAS}}
      && 59.2$\pm$0.4 & 56.0$\pm$0.8 \\
      &\checkmark& 60.5$\pm$0.2 & 58.1$\pm$0.8 \\
      COT \cite{COT} && 57.6$\pm$0.2 & 51.6$\pm$0.8\\
      \midrule
      \multirow{2}{*}{Ours}
      && 60.6$\pm$0.4 & 56.2$\pm$0.6\\
      &\checkmark& \textbf{62.1}$\pm$0.4 & \textbf{59.3}$\pm$0.3\\
      \bottomrule
      \end{tabular}
      }
\end{table}

\begin{table}[htbp]
  \caption{Classification accuracy (\%) averaged over 3 seeds ($\pm$ SEM) on the Sim-to-Real dataset. M11: ModelNet-11; SO*11: ScanObjectNN-11; S9: ShapeNet-9; SO*9: ScanObjectNN-9. Our method achieves best performance on the two adaptation scenarios.}
  \label{tab:sim2real}
  \centering
  \resizebox{0.9\linewidth}{!}{
      \begin{tabular}{lc||cc}
      \toprule
      Method & SPST & M11$\rightarrow$SO*11 & S9$\rightarrow$SO*9 \\
      \midrule
      w/o Adaptation && 61.68$\pm$1.26 & 57.42$\pm$1.01 \\
      \midrule
      PointDAN \cite{PointDAN} && 63.32$\pm$0.85 & 54.95$\pm$0.87 \\
      MetaSets \cite{MetaSets} && 72.42$\pm$0.21 & 60.92$\pm$0.76 \\
      GLRV \cite{GLRV} &\checkmark& 75.16$\pm$0.34 & 62.46$\pm$0.55 \\
      \midrule
      \multirow{2}{*}{Ours}
      &&  72.68$\pm$0.72 &  61.25$\pm$0.83\\
      &\checkmark&  \textbf{76.19}$\pm$0.47 &  \textbf{63.85}$\pm$0.57\\
      \bottomrule
      \end{tabular}
      }
\end{table}
\begin{table*}[htbp]
\centering
\caption{classification accuracy (\%) averaged over 3 seeds ($\pm$ SEM) on the GraspNetPC-10 dataset. Syn.: Synthetic domain, Kin.: Kinect domain, RS.: Realsense domain. Our models achieve the best performance over all settings. {$\dagger$} denotes experiments without using 3 seeds. * denotes the utilization of a more sophisticated self-training method.}
\label{tab:graspnet}
\resizebox{0.75\linewidth}{!}{
\begin{tabular}{lc||ccccc}
\toprule
Methods & SPST & Syn. $\rightarrow$ Kin. & Syn. $\rightarrow$ RS. & Kin. $\rightarrow$ RS. & RS. $\rightarrow$ Kin. & Avg. \\
\midrule
Oracle &  & 97.2$\pm$0.8 & 95.6$\pm$0.4 & 95.6$\pm$0.3 & 97.2$\pm$0.4 & 96.4 \\
w/o DA &  & 61.3$\pm$1.0 & 54.4$\pm$0.9 & 53.4$\pm$1.3 & 68.5$\pm$0.5 & 59.4 \\
\midrule
DANN \cite{dann} &  & 78.6$\pm$0.3 & 70.3$\pm$0.5 & 46.1$\pm$2.2 & 67.9$\pm$0.3 & 65.7 \\
PointDAN \cite{PointDAN} & & 77.0$\pm$0.2 & 72.5$\pm$0.3 & 65.9$\pm$1.2 & 82.3$\pm$0.5 & 74.4 \\
RS \cite{SSL_PointSeg_PC} & & 67.3$\pm$0.4 & 58.6$\pm$0.8 & 55.7$\pm$1.5 & 69.6$\pm$0.4 & 62.8 \\
DefRec+PCM \cite{PCM_RegRecT} & & 80.7$\pm$0.1 & 70.5$\pm$0.4 & 65.1$\pm$0.3 & 77.7$\pm$1.2 & 73.5 \\
GAST \cite{GAST} & & 69.8$\pm$0.4 & 61.3$\pm$0.3 & 58.7$\pm$1.0 & 70.6$\pm$0.3 & 65.1 \\
GAI \cite{GAI} & & 81.2$\pm$0.3 & 73.1$\pm$0.2 & 66.4$\pm$0.5 & 82.6$\pm$0.4 & 75.8\\
DAS{$\dagger$} \cite{DAS} & & 91.6$\pm$0.0 & 74.2$\pm$0.0 & 71.9$\pm$0.0 & 85.0$\pm$0.0 & 80.7 \\
Ours & & \textbf{97.2$\pm$0.5} & \textbf{84.1$\pm$0.6} & \textbf{79.9$\pm$1.4} & \textbf{92.3$\pm$0.8} & \textbf{88.4} \\
\midrule
GAST \cite{GAST} & $\checkmark$ & 81.3$\pm$1.8 & 72.3$\pm$0.8 & 61.3$\pm$0.9 & 80.1$\pm$0.5 & 73.8 \\
GAI \cite{GAI} &  $\checkmark$ & 94.6$\pm$0.4 & 80.5$\pm$0.2 & 76.8$\pm$0.4 & 85.9$\pm$0.3 & 84.4 \\
DAS \cite{DAS} &  $\checkmark$ & 97.0$\pm$0.2 & 79.6$\pm$0.8 & 79.1$\pm$0.7 & 95.7$\pm$0.7 & 87.8 \\
DAS* \cite{DAS} &  $\checkmark$ & 97.2$\pm$0.1 & 84.4$\pm$1.6 & 79.9$\pm$0.4 & 97.0$\pm$0.7 & 89.6 \\
Ours* &  $\checkmark$ & \textbf{97.8$\pm$0.7} & \textbf{84.9$\pm$0.6} & \textbf{82.6$\pm$0.8} & \textbf{98.2$\pm$1.2} & \textbf{90.9} \\
\bottomrule
\end{tabular}
}
\end{table*}

\begin{table*}[ht]
  \caption{Ablation study on each component of our method. Experiments are conducted on PointDA-10 dataset with the M10 $\rightarrow$ S*10 and the S10 $\rightarrow$ S*10 scenarios and on Sim-to-Real dataset with the M11 $\rightarrow$ SO*11 and the S9 $\rightarrow$ SO*9 scenarios. CDMix: Cross-Domain Mixup, SSL: Self-Supervised Learning Local Geometric Implicits, CLST: Contrastive Learning Self-Training.}
  \label{tab:ablation}
  \centering
  \resizebox{0.8\linewidth}{!}{
      \begin{tabular}{lccc||cccc}
      \toprule
      Method & CDMix & SSL & CLST & M10$\rightarrow$S*10 & S10$\rightarrow$S*10 & M11$\rightarrow$SO*11 & S9$\rightarrow$SO*9 \\
      \midrule
      w/o Adaptation & & & & 43.8$\pm$2.3 & 42.5$\pm$1.4 & 61.68$\pm$1.26 & 57.42$\pm$1.01 \\
      \midrule
      \multirow{4}{*}{Ours}
      & \checkmark & & & 57.7$\pm$0.4 & 54.8$\pm$0.5 & 66.95$\pm$0.86 & 57.50$\pm$0.73\\
      & & \checkmark & & 58.6$\pm$0.3 & 55.2$\pm$0.4 & 69.05$\pm$0.78 & 57.75$\pm$0.69\\
      & \checkmark & \checkmark & & 60.6$\pm$0.4 & 56.2$\pm$0.6 & 72.68$\pm$0.72 & 61.25$\pm$0.83\\
      & \checkmark & \checkmark & \checkmark &  \textbf{62.1}$\pm$0.4 & \textbf{59.3}$\pm$0.3 & \textbf{76.19}$\pm$0.47 &  \textbf{63.85}$\pm$0.57\\
      \bottomrule
      \end{tabular}
      }
\end{table*}
\begin{table}[htbp]
  \caption{Classification accuracy (\%) averaged over 3 seeds ($\pm$ SEM) on the PointDA-10 dataset. M10: ModelNet-10; S10: ShapeNet-10; S*10: ScanNet-10. Our method achieves comparable performance to state-of-the-art methods in 4 non-Sim2Real scenarios.}
  \label{tab:pointda_ns2r}
  \centering
  \resizebox{0.98\linewidth}{!}{
      \begin{tabular}{l||cccc}
      \toprule
      Method & M10$\rightarrow$S10 & S10$\rightarrow$M10 & S*10$\rightarrow$M10 & S*10$\rightarrow$S10\\
      \midrule
      w/o Adaptation & 83.3$\pm$0.7 & 75.5$\pm$1.8 & 63.8$\pm$3.9 & 64.2$\pm$0.8\\
      \midrule
      DANN \cite{dann} & 74.8$\pm$2.8 & 57.5$\pm$0.4 & 43.7$\pm$2.9 & 71.6$\pm$1.0\\
      PointDAN \cite{PointDAN} & 83.9$\pm$0.3 & 63.3$\pm$1.1 & 43.6$\pm$2.0 & 56.4$\pm$1.5\\
      RS \cite{SSL_PointSeg_PC} & 79.9$\pm$0.8 & 75.2$\pm$2.0 & 71.8$\pm$2.3 & 71.2$\pm$2.8\\
      DefRec \cite{PCM_RegRecT} & 82.7$\pm$0.6 & 79.8$\pm$0.5 & 66.0$\pm$0.8 & 67.4$\pm$1.2\\
      DefRec+PCM \cite{PCM_RegRecT} & 81.7$\pm$0.6 & 78.6$\pm$0.7 & 73.7$\pm$1.6 & 71.1$\pm$1.4\\
      GAST \cite{GAST} & 84.8$\pm$0.1 & 80.8$\pm$0.6 & 81.1$\pm$0.8 & 74.9$\pm$0.5\\
      GAI \cite{GAI} &86.2$\pm$0.2 & 81.4$\pm$0.4 & 81.5$\pm$0.5 & 74.4$\pm$0.4\\
      GLRV \cite{GLRV} & 85.4$\pm$0.4 & 78.8$\pm$0.6 & 77.8$\pm$1.1 & 76.2$\pm$0.6\\
      PDG \cite{PDG} & 85.6$\pm$0.3 & 73.1$\pm$0.4 & 70.3$\pm$1.3 & 66.3$\pm$0.8\\
      SD$^\dagger$ \cite{SD} & 83.9$\pm$0.0 & 80.3$\pm$0.0 & 85.5$\pm$0.0 & 80.9$\pm$0.0\\
      DAS \cite{DAS} & \textbf{87.2}$\pm$0.9 & 82.4$\pm$0.7 & 84.8$\pm$2.3 & 82.3$\pm$1.5\\
      COT \cite{COT} & 84.7$\pm$0.2 & \textbf{89.6}$\pm1.4$ & 85.5$\pm2.2$ & 77.6$\pm$0.5\\
      \midrule
      Ours  & 86.4$\pm$0.3 & 82.9$\pm$0.3 & \textbf{85.6}$\pm$0.7 & \textbf{82.6}$\pm$0.5\\
      \bottomrule
      \end{tabular}
      }
\end{table}

\begin{table}[ht]
  \caption{Ablation study on complementarity of the fourier positional encoder. Experiments are conducted on two Sim2Real UDA benchmarks. "+" indicates the network of the global feature encoder $\Phi_{g}$.}
  \label{tab:ablation_pointNN}
  \centering
  \resizebox{0.98\linewidth}{!}{
      \begin{tabular}{l||cccc}
      \toprule
      Method & M10$\rightarrow$S*10 & S10$\rightarrow$S*10 & M11$\rightarrow$SO*11 & S9$\rightarrow$SO*9 \\
      \midrule
       
       + DGCNN \cite{DGCNN} & 59.2$\pm$0.3 & 58.3$\pm$0.6 & 73.45$\pm$0.64 & 60.35$\pm$0.52\\
       + PointNet \cite{Pointnet} & 56.6$\pm$0.5 & 52.7$\pm$0.6 & 66.65$\pm$0.52 & 56.95$\pm$0.63\\
       + Point-NN \cite{Point_NN} & 49.6$\pm$0.5 & 47.7$\pm$0.7 & 58.45$\pm$0.65 & 52.95$\pm$0.63\\
       + Point-PN \cite{Point_NN} & \textbf{62.1}$\pm$0.4 & \textbf{59.3}$\pm$0.3 & \textbf{76.19}$\pm$0.47 &  \textbf{63.85}$\pm$0.57\\
     
      \bottomrule
      \end{tabular}
      }
\end{table}

\subsection{Comparsion to the State-of-the-art}
For the \textbf{PointDA-10} dataset, we compare our method with the state-of-the-art point-based domain adaptation methods, including Domain Adversarial Neural Network (DANN) \cite{dann}, Point Domain Adaptation Network (PointDAN) \cite{PointDAN}, Reconstruction Space Network (RS) \cite{SSL_PointSeg_PC}, Deformation Reconstruction Network with Point Cloud Mixup (DefRec + PCM) \cite{PCM_RegRecT}, Geometry-Aware Self-Training (GAST) \cite{GAST}, Geometry-Aware Implicits (GAI) \cite{GAI}, Global-Local Reliable Voting (GLRV) \cite{GLRV}, Part-based Domain Generalization (PDG) \cite{PDG}, Self-Distillation (SD) \cite{SD}, Domain Adaptative Sampling (DAS) \cite{DAS} and  Cross-domain Optimal Transport (COT) \cite{COT}.
We report the mean accuracy and standard error with three seeds in Table \ref{tab:pointda}. Our method achieves the best accuracy on Sim2Real unsupervised domain adaptation scenarios. For the sake of fairness, we also report the performance under four other non-Sim2Real scenarios. As shown in Table \ref{tab:pointda_ns2r}, 
our method achieves comparable performance to state-of-the-art methods in 4 non-Sim2Real scenarios. For all 6 settings, our TAM achieves an average accuracy of 76.5\%, which is 0.6\% higher than the current state-of-the-art method DAS \cite{DAS}.

For the \textbf{Sim-to-Real} dataset, we compare our method with the meta-learning approach MetaSets \cite{MetaSets}, as well as point-based domain adaptation methods like PointDAN \cite{PointDAN} and GLRV \cite{GLRV}. Table \ref{tab:sim2real}
reports the mean accuracy and standard error across three seeds. Our method surpasses both point-based domain adaptation and meta-learning approaches, setting a new state-of-the-art.

For the \textbf{GraspNetPC-10} dataset, we compare our method with DANN \cite{dann}, PointDAN \cite{PointDAN}, RS \cite{SSL_PointSeg_PC}, DefRec + PCM \cite{PCM_RegRecT}, GAST \cite{GAST}, GAI \cite{GAI} and DAS \cite{DAS}. As presented in Table \ref{tab:graspnet}, our method achieved the best performance, highlighting the significance of both global and local information in adapting to diverse conditions. This result demonstrates that the local geometric information captured by our TAM model is essential for ensuring robust performance, thus confirming its importance within our theoretical framework.

\subsection{Ablation Study}
\noindent \textbf{Influence of CDMix and Local Implicit Fields --}
As shown in Table \ref{tab:ablation}, both the proposed CDMix and the local implicit representation components play a positive role in improving the domain adaptation in Sim2Real scenarios. The domain adaptation performance of using CDMix or the local implicit representation component alone is not very prominent, but the combination of the two works well, proving the complementary properties of the two modules. Note that we can achieve comparable performance to the state-of-the-art methods even without using CLST, which shows the effectiveness of the proposed method.

\begin{figure}[t]
\centering
\subfigure[$\mathcal{A}$-distance]{
\label{fig:A_Dist}
\begin{minipage}[b]{0.3\textwidth}
\includegraphics[width=1\textwidth, height=\textwidth]{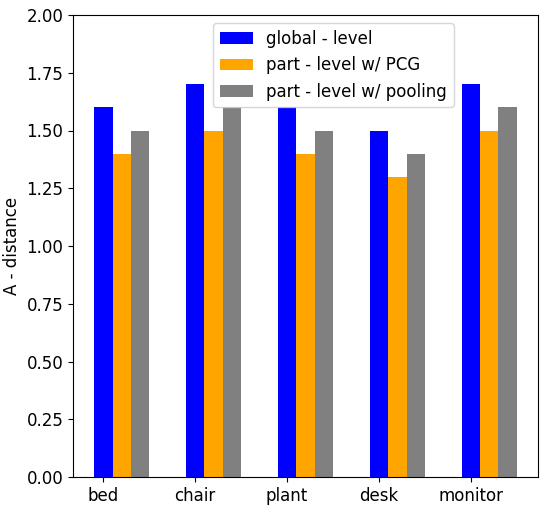}
\end{minipage}}
\subfigure[Discrimination]{
\label{fig:Dicrim}
\begin{minipage}[b]{0.16\textwidth}
\includegraphics[width=\textwidth, height=1.85\textwidth]{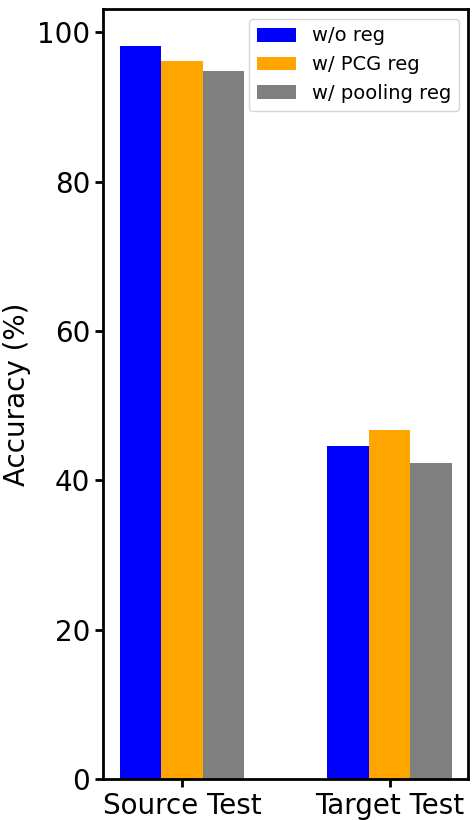}
\end{minipage}}
\caption{(a) $\mathcal{A}$-distance of different aggregated features between source and target domain. (b) Discrimination ability of global-level features regularized by different aggregated part-level feature measured by classification accuracy on test data of source and target domain.}
\label{fig:part_generation}
\vspace{-0.5cm}
\end{figure}

\noindent \textbf{Influence of PCG module --}
\cite{PDG} has demonstrated that part-level features are more generalizable but less discriminative than global-level features. Following \cite{PDG}, we also use $\mathcal{A}$-distance as a measure to evaluate the distribution discrepancy \cite{da_theory2, da_theory4}. It is defined as $dist_{\mathcal{A}}= 2(1 - \eta)$, where $\eta$ is the test error of a classifier trained to discriminate the source and the target domain data. In this work, the topologically aggregated features of the proposed PCG module can be regarded as a regularization of global features. Please refer to Equation \ref{eq:pcg_sim_loss}.
To analyze its influence, we train a feature encoder $\Phi_{fea}$, specifically a DGCNN \cite{DGCNN}, without utilizing high-frequency global spatial information on the source domain \textbf{M10} (ModelNet \cite{Modelnet}) and test it on the target domain \textbf{S*10} (ScanNet \cite{Scannet}).
Taking a point cloud sample $\mathcal{P}$ as input, we obtain a global-level feature $\bm{z} \in \mathcal{Z}$. As described in Section \ref{PCG}, by setting different $N$ and $k$, we can obtain local-level features at different scales, \ie 8 parts with 512 points in each part ($N=8, k=512$) for $\bm{z}^8 \in \{\bm{z}_i^{p}\}_{i=1}^8$ and 16 parts with 256 points in each part ($N=16, k=256$) for $\bm{z}^{16} \in \{\bm{z}_i^{p}\}_{i=1}^{16}$.  For illustration, we choose the case of 16 parts with 256 points per part. As shown in Fig. \ref{fig:A_Dist}, the $dist_\mathcal{A}$ values of part-level features with different aggregation strategies (\ie PCG or pooling) are smaller than those of global-level features in most categories, indicating that part-level features generalize better than global-level features. This result is consistent with the observation of \cite{PDG}. We also train a classifier $\Phi_{cls}$, \ie MLP, on source domain training data and test on source domain test data and target test data to evaluate the discrimination ability of global-level features regularized by different aggregated part-level features. As shown in Fig. \ref{fig:Dicrim}, the discrimination abilities of the global features regularized by the topologically aggregated features of the PCG module perform better. 

\noindent \textbf{Influence of Fourier Positional Encoding --}
\cite{Point_NN, fourier, nerf} demonstrated that Fourier Positional Encoding, using trigonometric functions to encode raw point cloud coordinates, effectively captures rich high-frequency 3D spatial information. This method complements traditional point cloud deep networks like DGCNN \cite{DGCNN}, PointNet \cite{Pointnet}, PointNet++ \cite{Pointnet++}, and PCT \cite{PCT}, enhancing classification performance by combining low-level high-frequency spatial structures with high-level semantic features.
To further explore the cross-domain generalization and robustness of high-frequency 3D spatial information, we follow \cite{Point_NN} by incorporating trigonometric functions as a plug-and-play module to capture this structure. As shown in Table \ref{tab:ablation_pointNN}, traditional point cloud networks that lack these functions perform significantly worse on Sim2Real benchmarks. We also compare our method with Point-NN \cite{Point_NN}, which uses trigonometric encoding but lacks sufficient parameters to capture robust semantic information. We added a trainable projector and classifier, yet the limited parameters in Point-NN still result in suboptimal performance.
These experiments demonstrate that high-frequency 3D spatial structures exhibit strong cross-domain invariance, and when combined with global semantic information, they significantly boost cross-domain generalization and robustness.

\begin{figure}[t]
  \centering
   \includegraphics[width=0.95\linewidth]{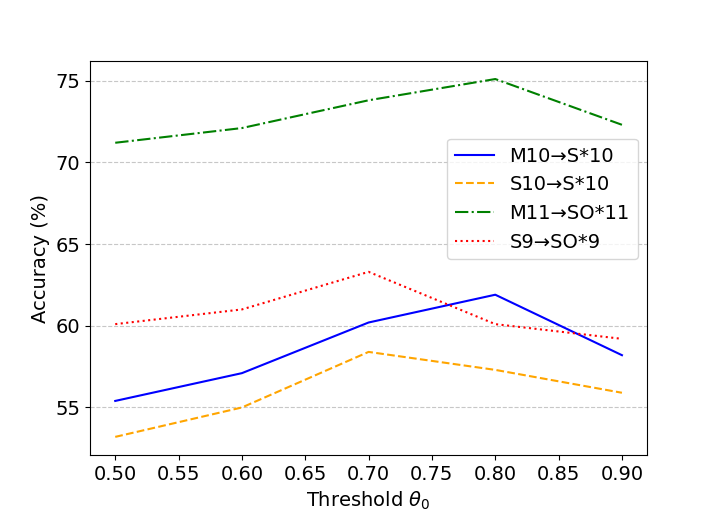}
   \caption{Illustration of the effect of prediction confidence threshold $\theta_0$ in self-training stage.}
   \label{fig:threshold}
\end{figure}

\noindent \textbf{Influence of confidence threshold $\theta$ in self-training --}
During self-training, the choice of threshold has a great impact on performance. A threshold that is too large will result in insufficient pseudo-label samples, while a threshold that is too small will result in the introduction of too many noise samples. Therefore, to obtain a suitable threshold, we conducted multiple experiments with different thresholds on four Sim2Real settings. As shown in Fig. \ref{fig:threshold}, we found that when the threshold $\theta$ is close to 0.8, \textbf{M10} $\rightarrow$ \textbf{S*10} and \textbf{M11} $\rightarrow$ \textbf{SO*11} achieve the best performance, and when the threshold $\theta$ is close to 0.7, \textbf{S10} $\rightarrow$ \textbf{S*10} and \textbf{S9} $\rightarrow$ \textbf{SO*9} achieve the best performance.

\subsection{Visualization}
\begin{figure}[ht]
    \centering
    \subfigure[\scriptsize PointDAN \cite{PointDAN}]{
    \begin{minipage}[b]{0.1\textwidth}
        \label{Fig.PointDAN}
            \includegraphics[width=\textwidth]{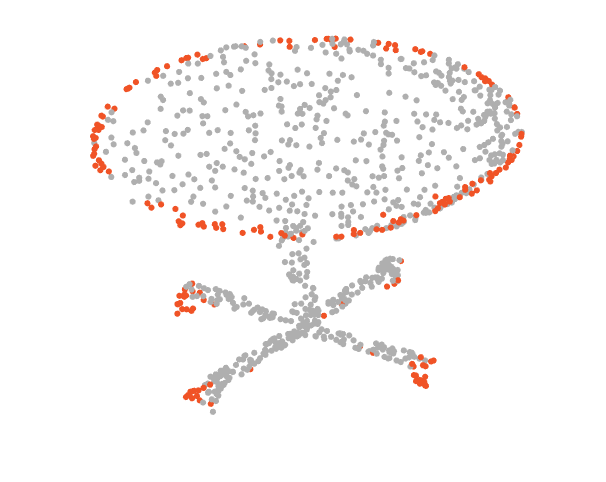}
            \includegraphics[width=\textwidth]{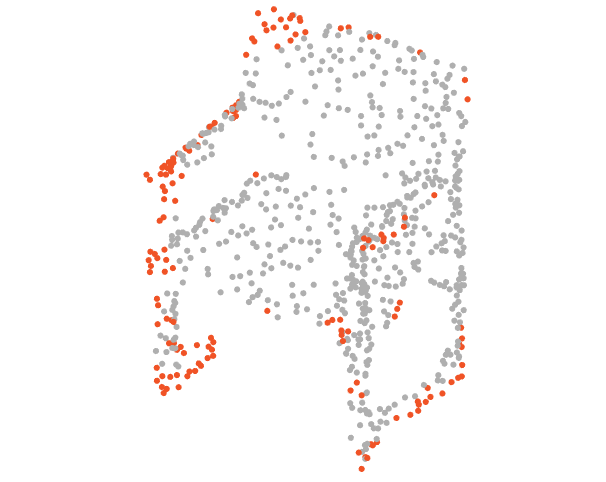}
            \includegraphics[width=\textwidth]{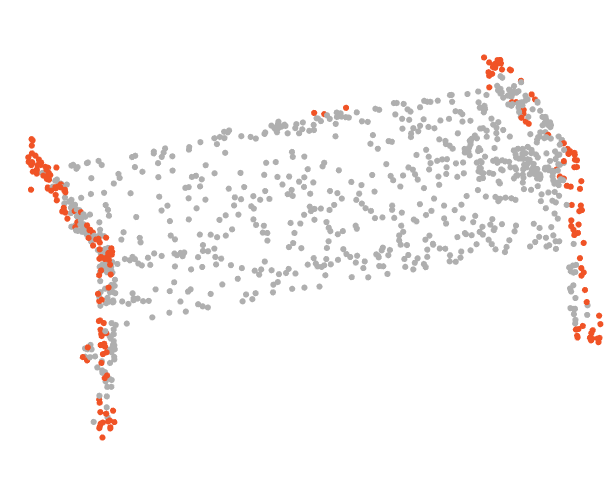}
            \includegraphics[width=\textwidth]{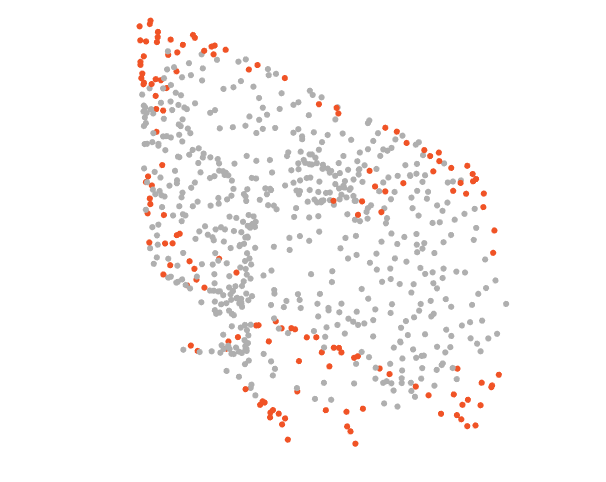}
    \end{minipage}
    }
    \subfigure[\scriptsize DefRec \cite{PCM_RegRecT}]{
    \begin{minipage}[b]{0.1\textwidth}
        \label{Fig.DefREC}
            \includegraphics[width=\textwidth]{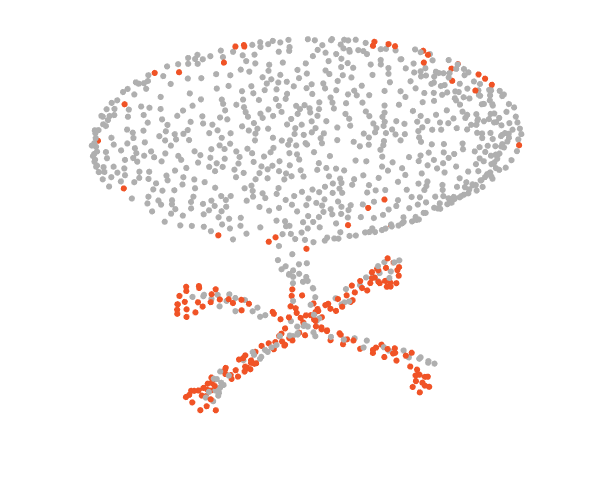}
            \includegraphics[width=\textwidth]{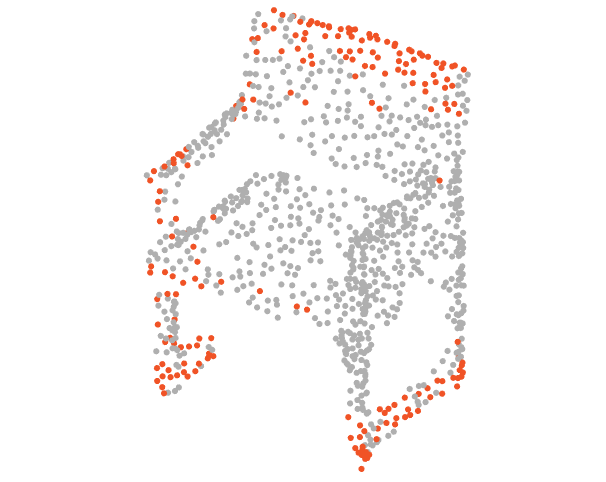}
            \includegraphics[width=\textwidth]{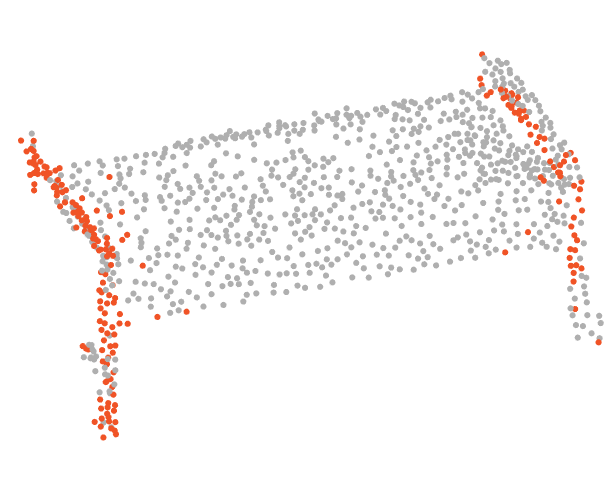}
            \includegraphics[width=\textwidth]{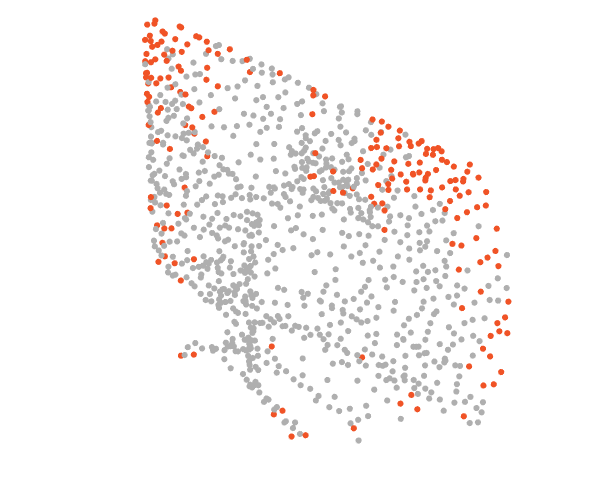}
    \end{minipage}
    }
    \subfigure[\scriptsize SD \cite{SD}]{
    \begin{minipage}[b]{0.1\textwidth}
        \label{Fig.SD}
            \includegraphics[width=\textwidth]{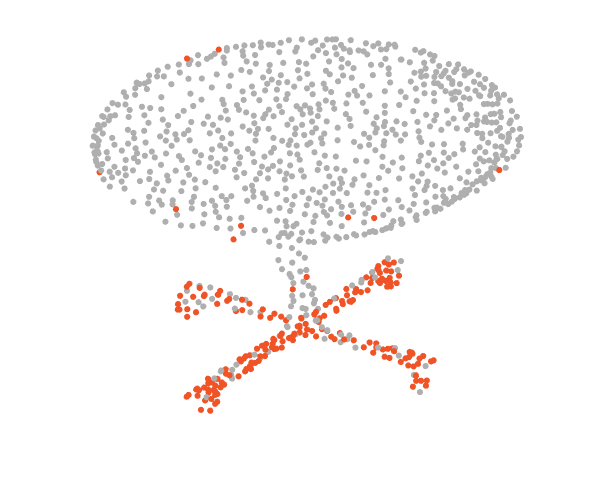}
            \includegraphics[width=\textwidth]{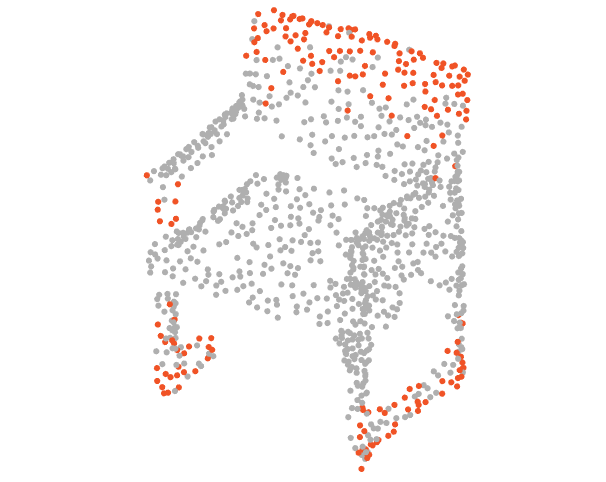}
            \includegraphics[width=\textwidth]{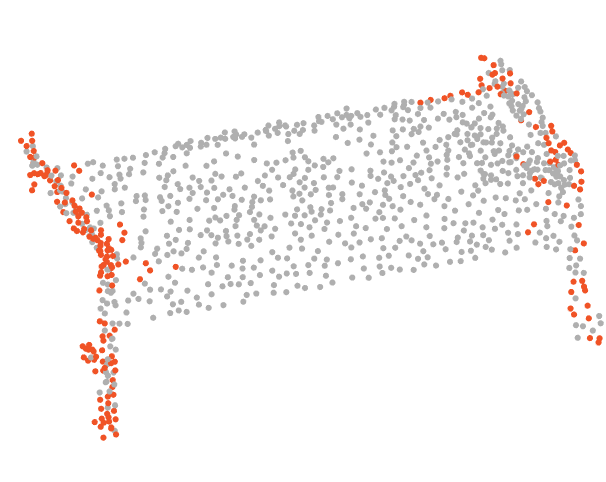}
            \includegraphics[width=\textwidth]{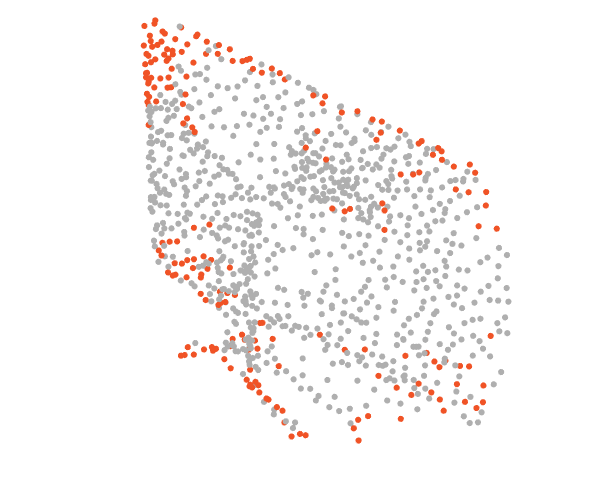}
    \end{minipage}
    }
    \subfigure[\scriptsize Ours]{
    \begin{minipage}[b]{0.1\textwidth}
        \label{Fig.TAM}
            \includegraphics[width=\textwidth]{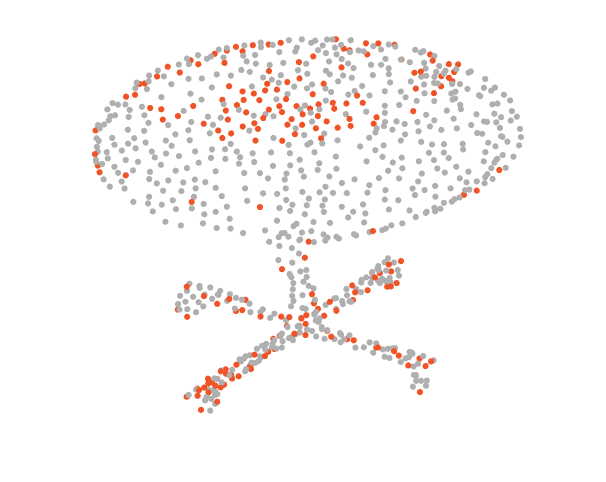}
            \includegraphics[width=\textwidth]{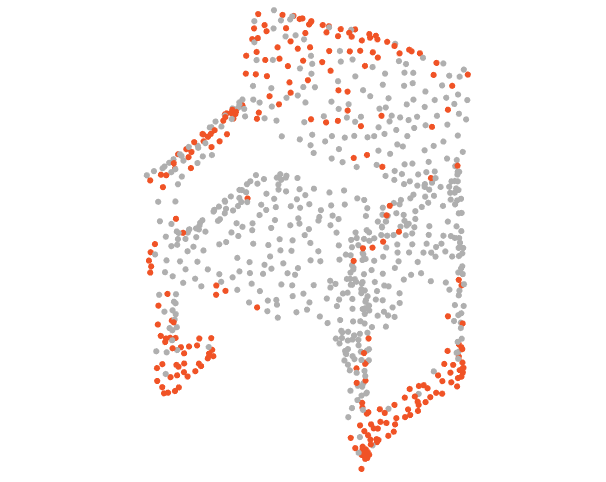}
            \includegraphics[width=\textwidth]{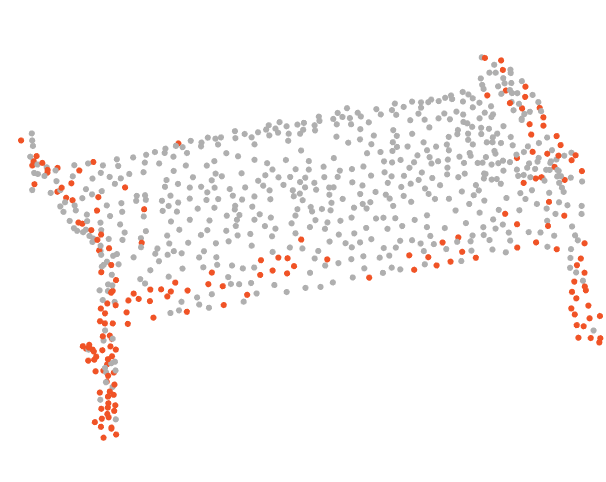}
            \includegraphics[width=\textwidth]{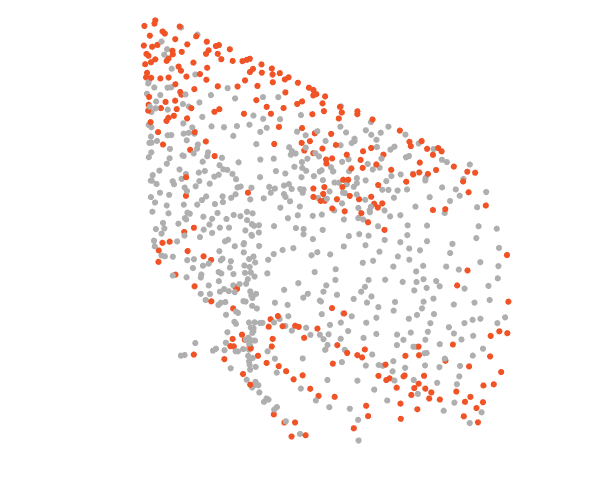}
    \end{minipage}
    }
    \caption{Saliency map visualization of various comparison methods under the setting of M$\rightarrow$S* on PointDA-10.} 
    \label{fig:saliency_map}
\end{figure}

We employ the point cloud saliency analysis technique cited in \cite{pc_saliency} to visualize the features of various comparison methods within the M$\rightarrow$S* setting on the PointDA-10 \cite{PointDAN} dataset. As depicted in Fig. \ref{fig:saliency_map}, our findings indicate that our method takes into account both global spatial topology and local geometric structures very well.
While placing emphasis on global features, our approach also devotes significant attention to local geometric nuances. This enables our method to extract semantic representations that are both robust and invariant. In contrast, a majority of other methods utilize DGCNN as the backbone for feature extraction. These methods tend to be overly preoccupied with local high-frequency areas, such as edges, which consequently impairs their generalization ability. 
In contrast, our method integrates local geometric information in a manner that enriches the global understanding, aligning seamlessly with the cognitive neuroscience perspectives on visual recognition.

\begin{figure*}[ht]
    \centering
    \subfigure[w/o Adapt: M10 $\rightarrow$ S*10]{
        \label{Fig.cofmt_m_sup.1}
        \begin{minipage}[b]{0.23\textwidth}
            \includegraphics[width=1\textwidth]{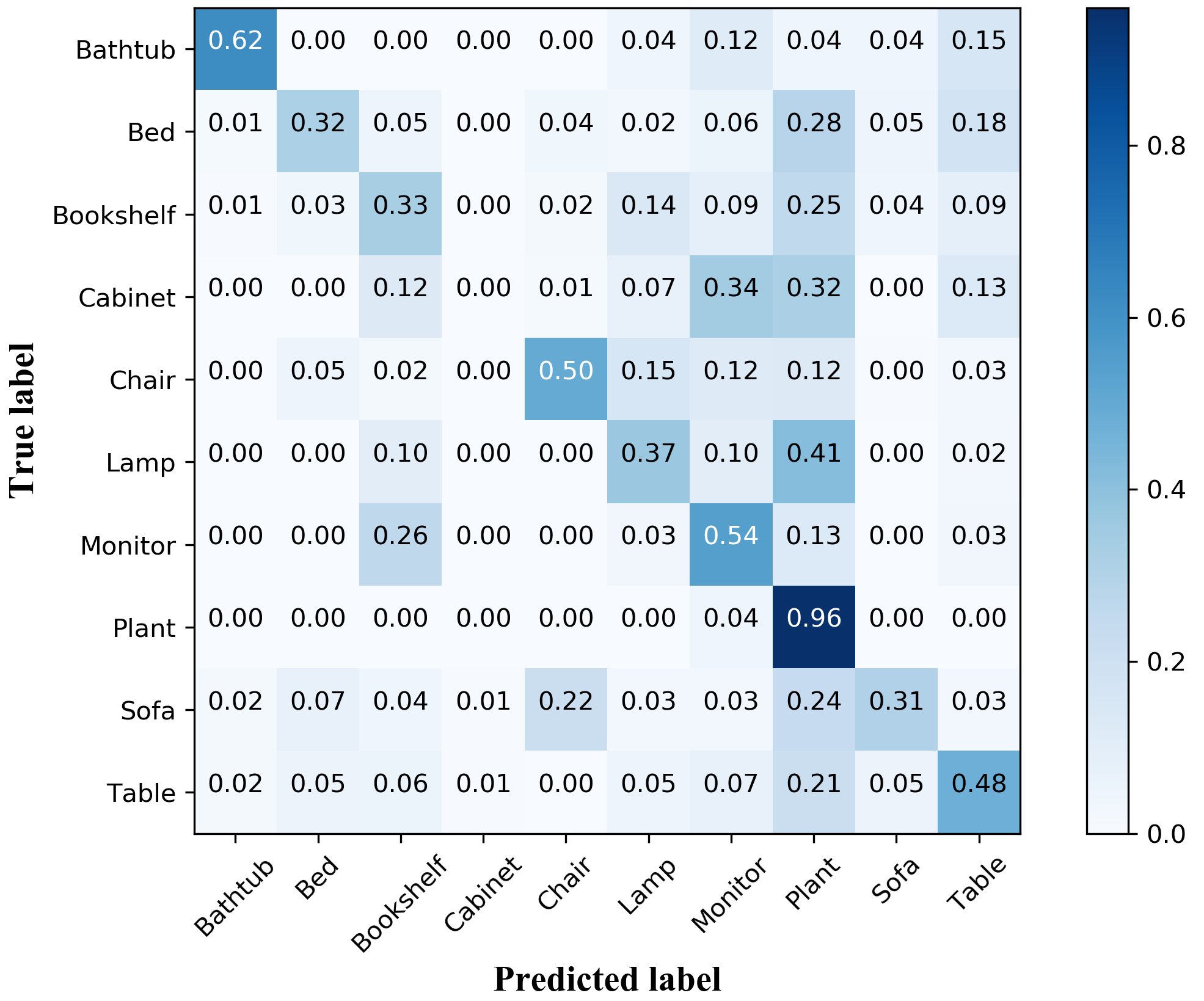}
        \end{minipage}}
    \subfigure[Ours: M10 $\rightarrow$ S*10]{
        \label{Fig.cofmt_m2r.2}
        \begin{minipage}[b]{0.23\textwidth}
            \includegraphics[width=1\textwidth]{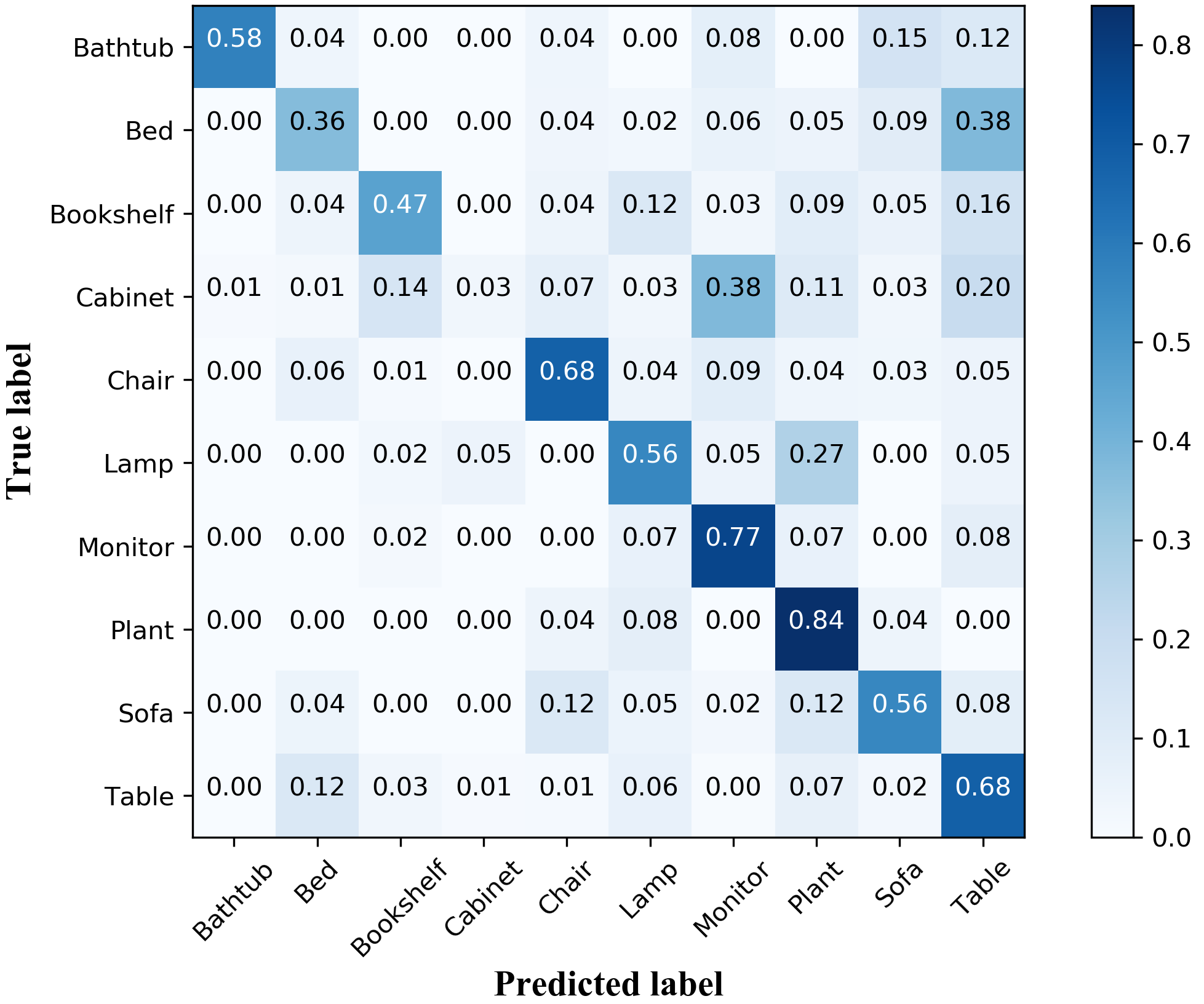}
        \end{minipage}}       
    \subfigure[w/o Adapt: S10 $\rightarrow$ S*10]{
        \label{Fig.cofmt_s_sup.3}
        \begin{minipage}[b]{0.23\textwidth}
            \includegraphics[width=1\textwidth]{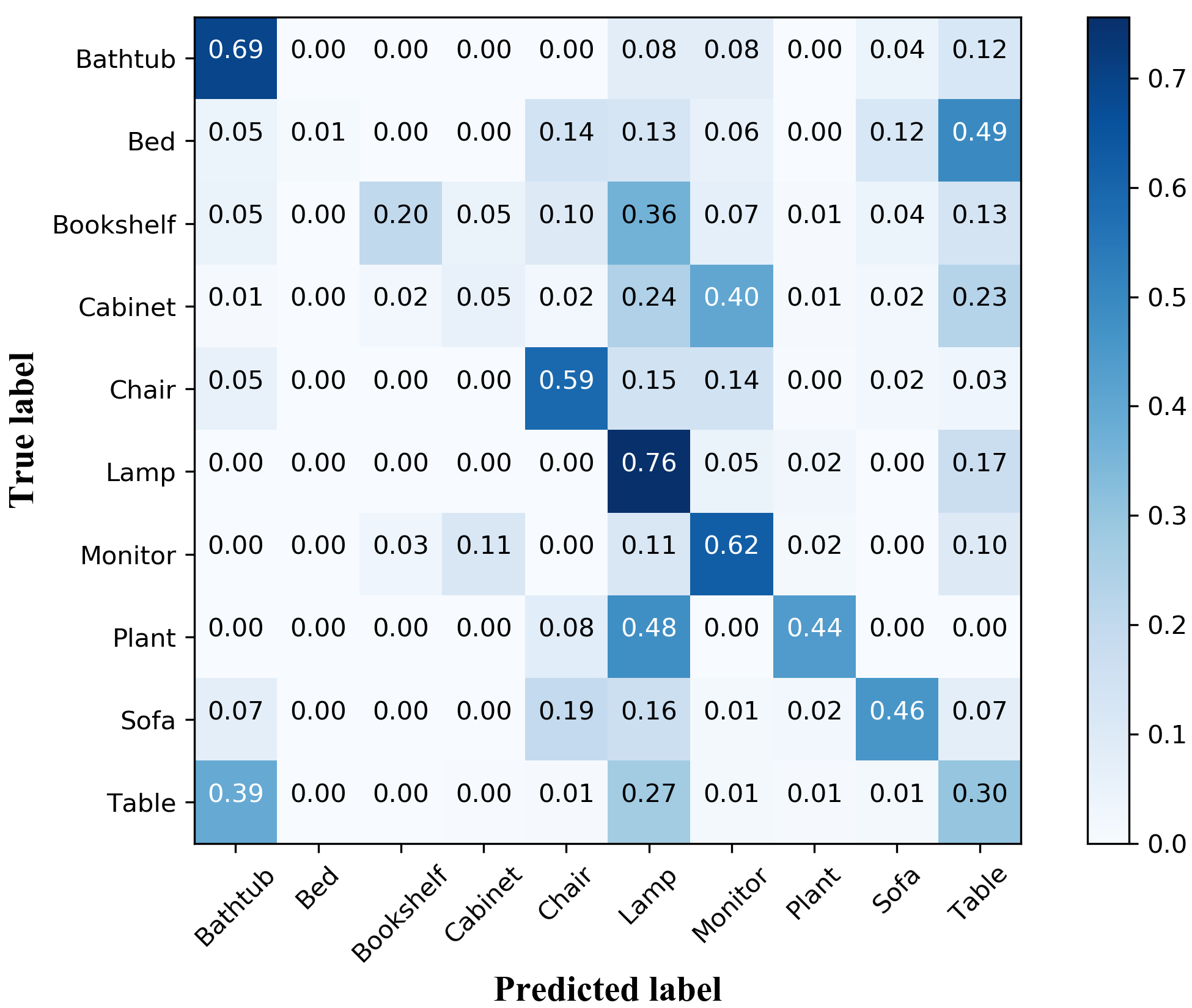}
        \end{minipage}}
    \subfigure[Ours: S10 $\rightarrow$ S*10]{
        \label{Fig.cofmt_s2r.4}
        \begin{minipage}[b]{0.23\textwidth}
            \includegraphics[width=1\textwidth]{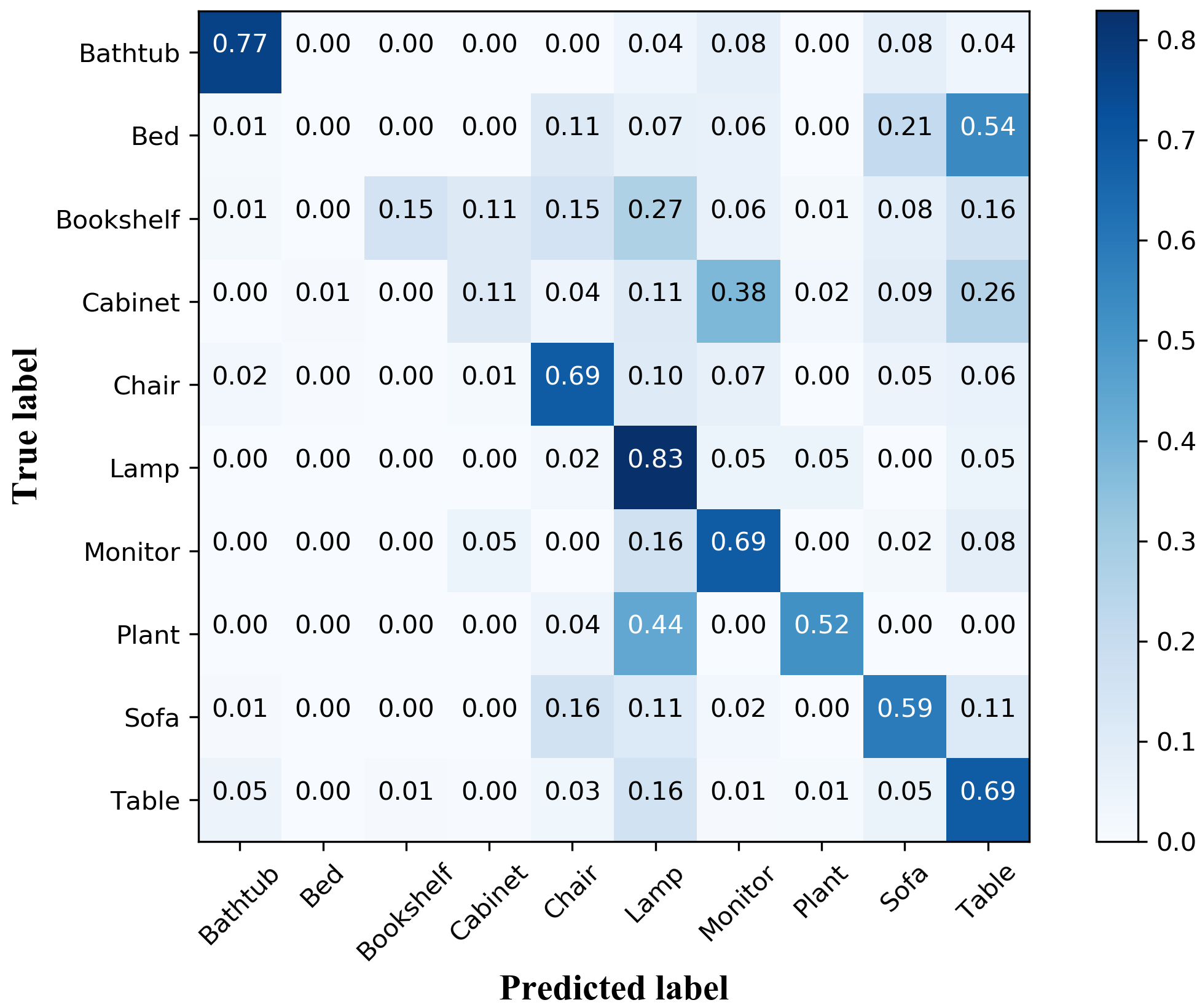}
        \end{minipage}}   
    \subfigure[w/o Adapt: M11 $\rightarrow$ SO*11]{
        \label{Fig.cofmt_m11_sup.1}
        \begin{minipage}[b]{0.23\textwidth}
            \includegraphics[width=1\textwidth]{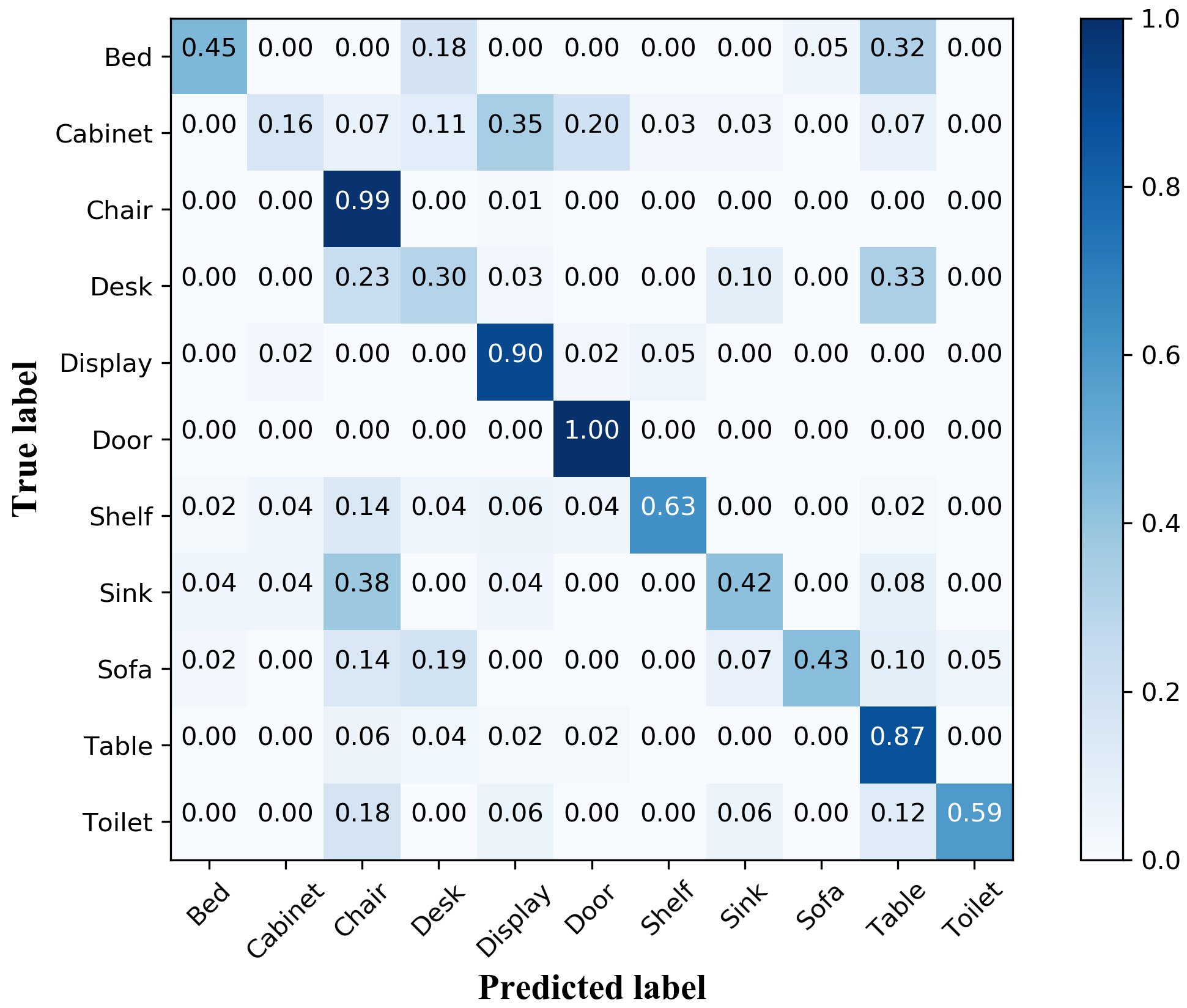}
        \end{minipage}}
    \subfigure[Ours: M11 $\rightarrow$ SO*11]{
        \label{Fig.cofmt_m112r11.2}
        \begin{minipage}[b]{0.23\textwidth}
            \includegraphics[width=1\textwidth]{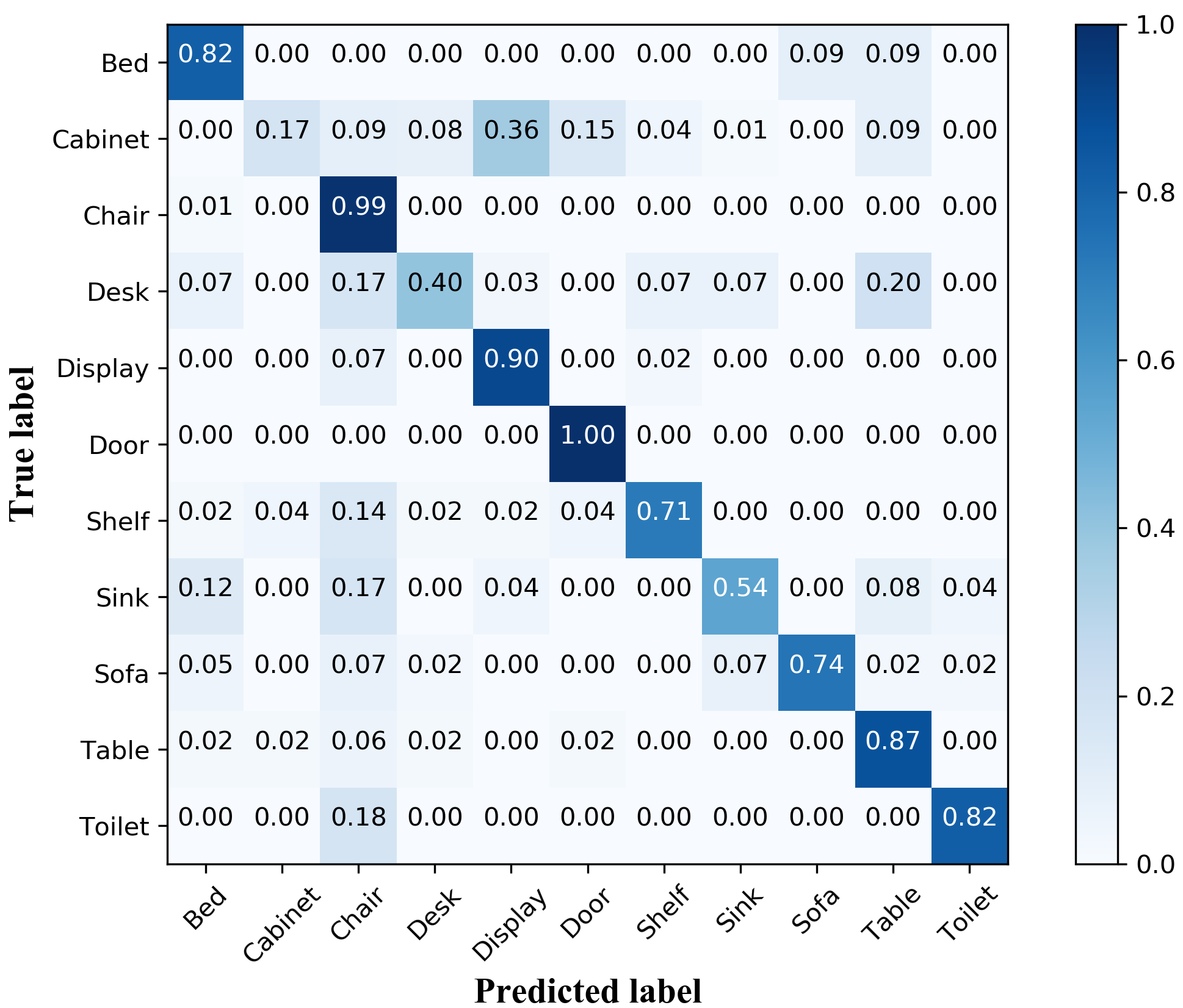}
        \end{minipage}}    
    \subfigure[w/o Adapt: S9 $\rightarrow$ SO*9]{
        \label{Fig.cofmt_s9_sup.3}
        \begin{minipage}[b]{0.23\textwidth}
            \includegraphics[width=1\textwidth]{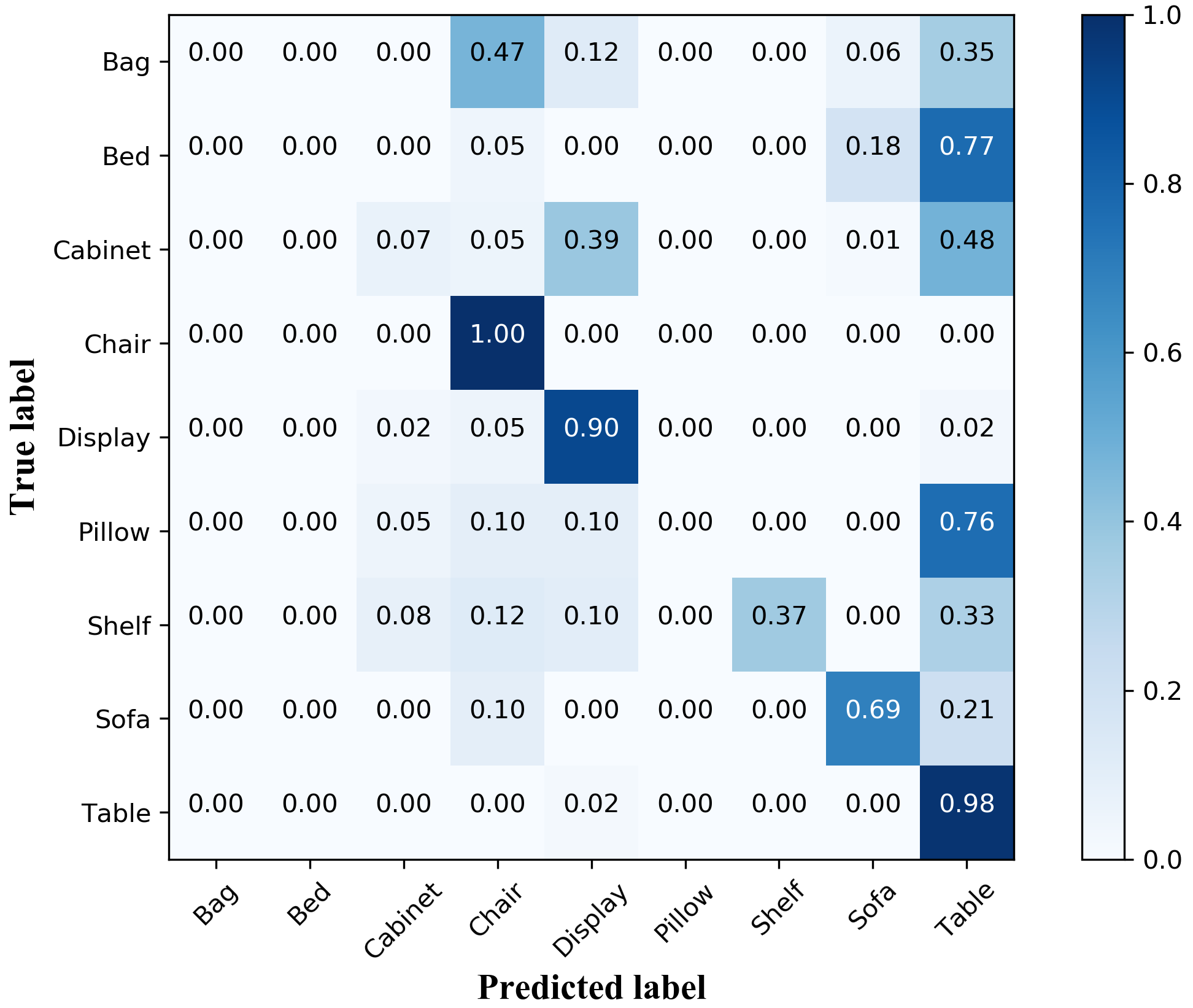}
        \end{minipage}}
    \subfigure[Ours: S9 $\rightarrow$ SO*9]{
        \label{Fig.cofmt_s92r9.4}
        \begin{minipage}[b]{0.23\textwidth}
            \includegraphics[width=1\textwidth]{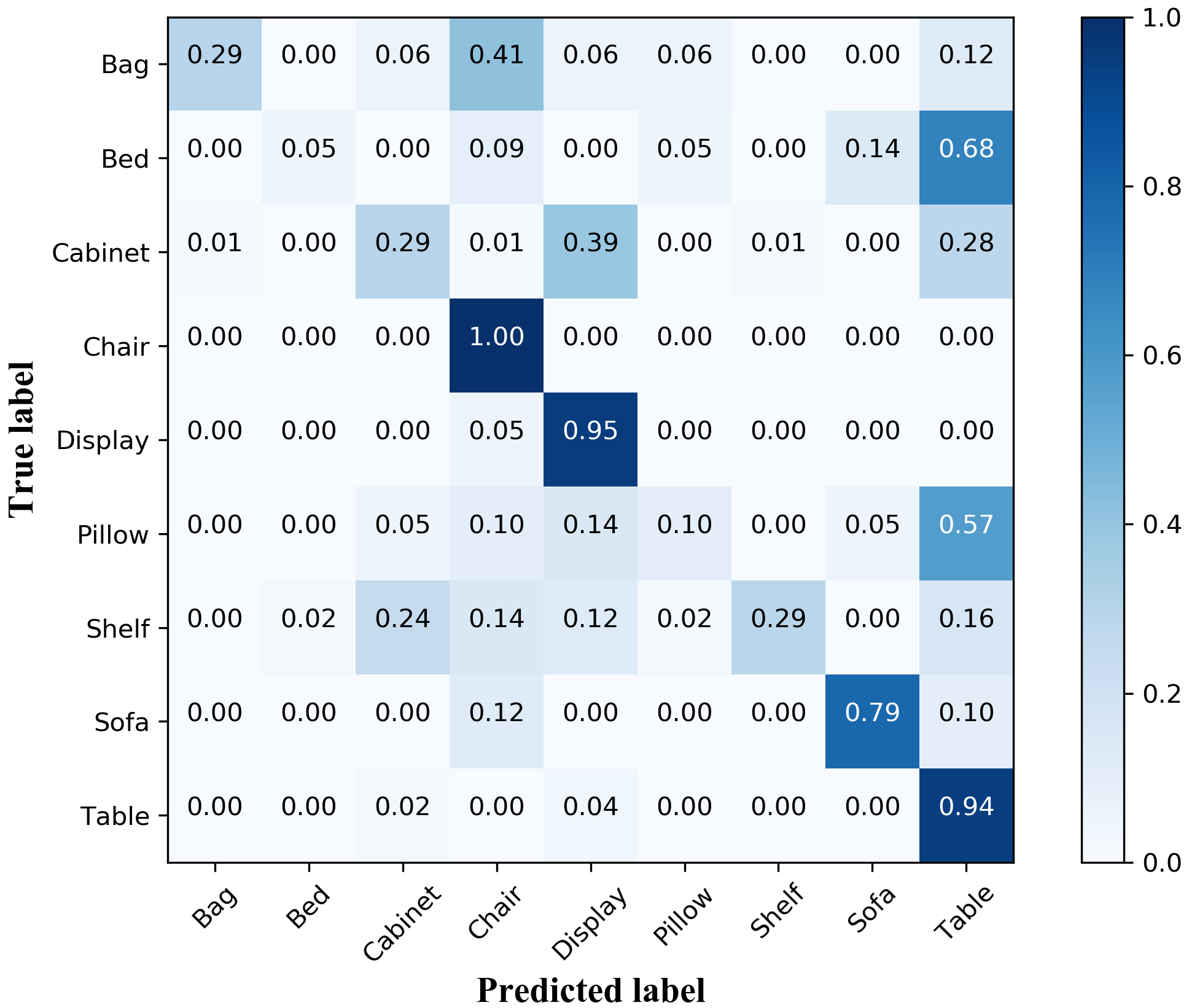}
        \end{minipage}}
    \caption{Confusion matrices of classifying testing samples on target domain.} 
    \label{Fig.confusion_matrix}
\end{figure*}

\begin{figure*}[ht]
    \centering
    \subfigure[w/o Adapt: M10 $\rightarrow$ S*10]{
        \label{fig:m2r1}
        \begin{minipage}[b]{0.23\textwidth}
            \includegraphics[width=1\textwidth]{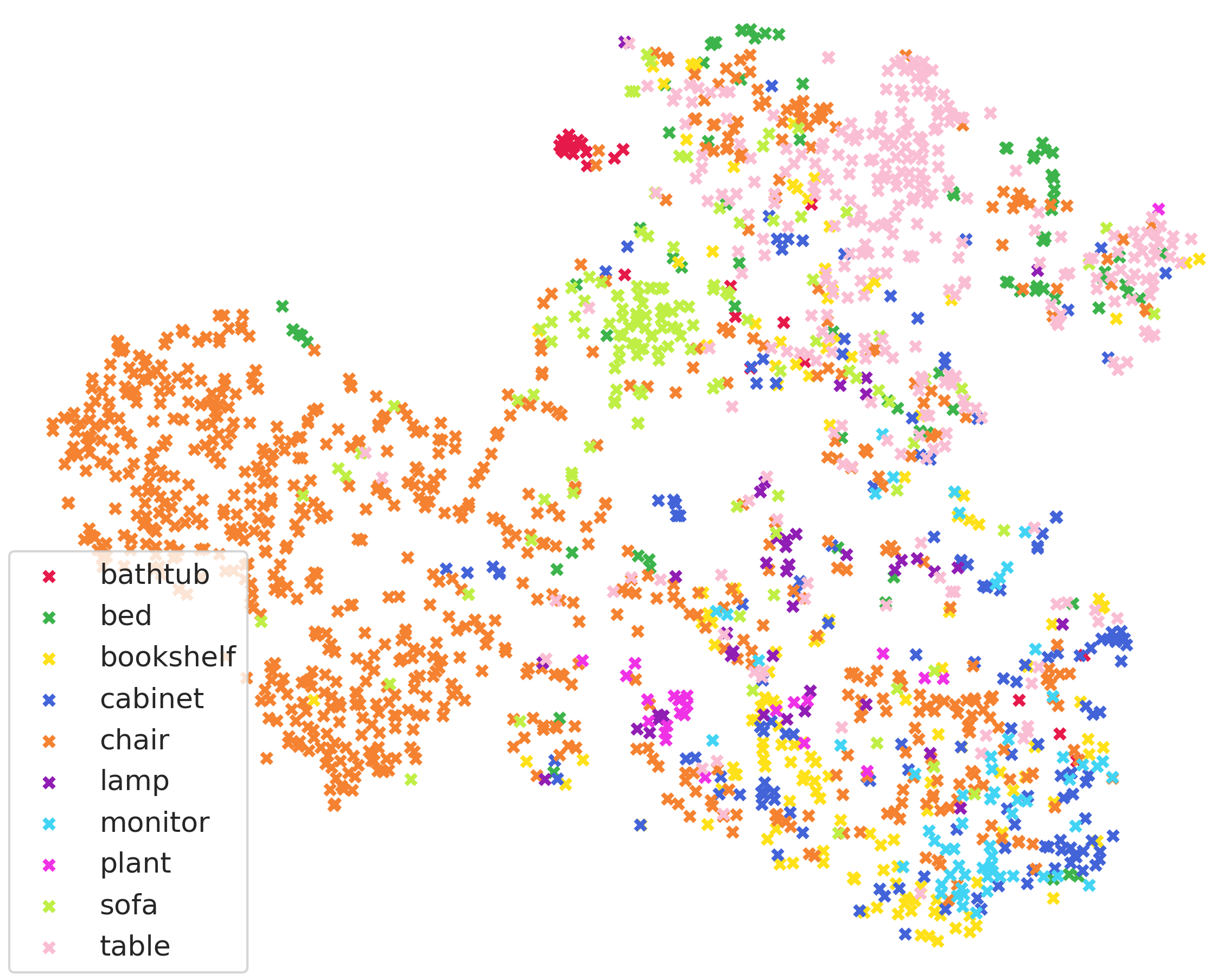}
        \end{minipage}}
    \subfigure[Ours: M10 $\rightarrow$ S*10]{
        \label{fig:m2r2}
        \begin{minipage}[b]{0.23\textwidth}
            \includegraphics[width=\textwidth]{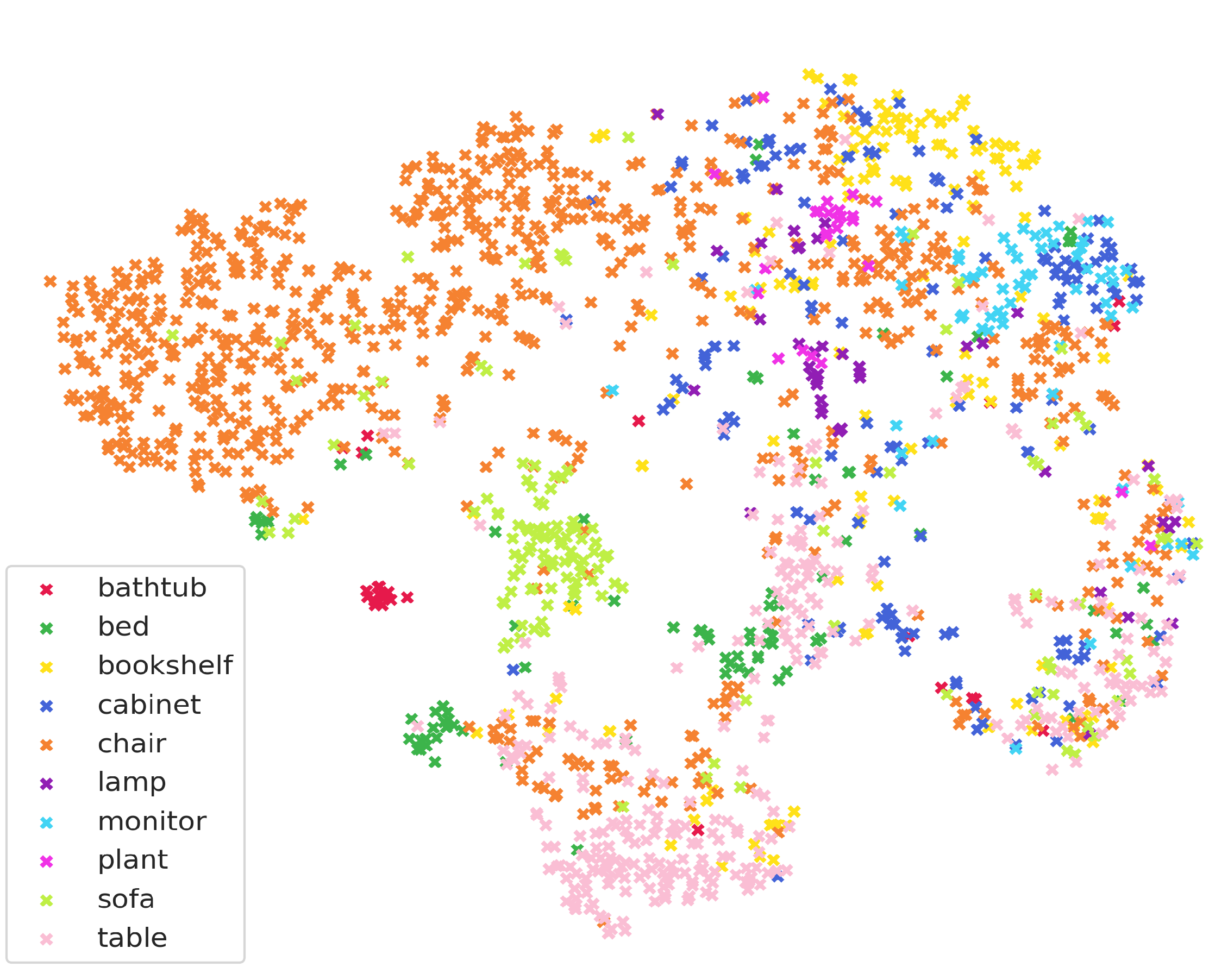}
    \end{minipage}}
    \subfigure[w/o Adapt: M11 $\rightarrow$ SO*11]{
        \label{fig:m2r13}
        \begin{minipage}[b]{0.23\textwidth}
            \includegraphics[width=1\textwidth]{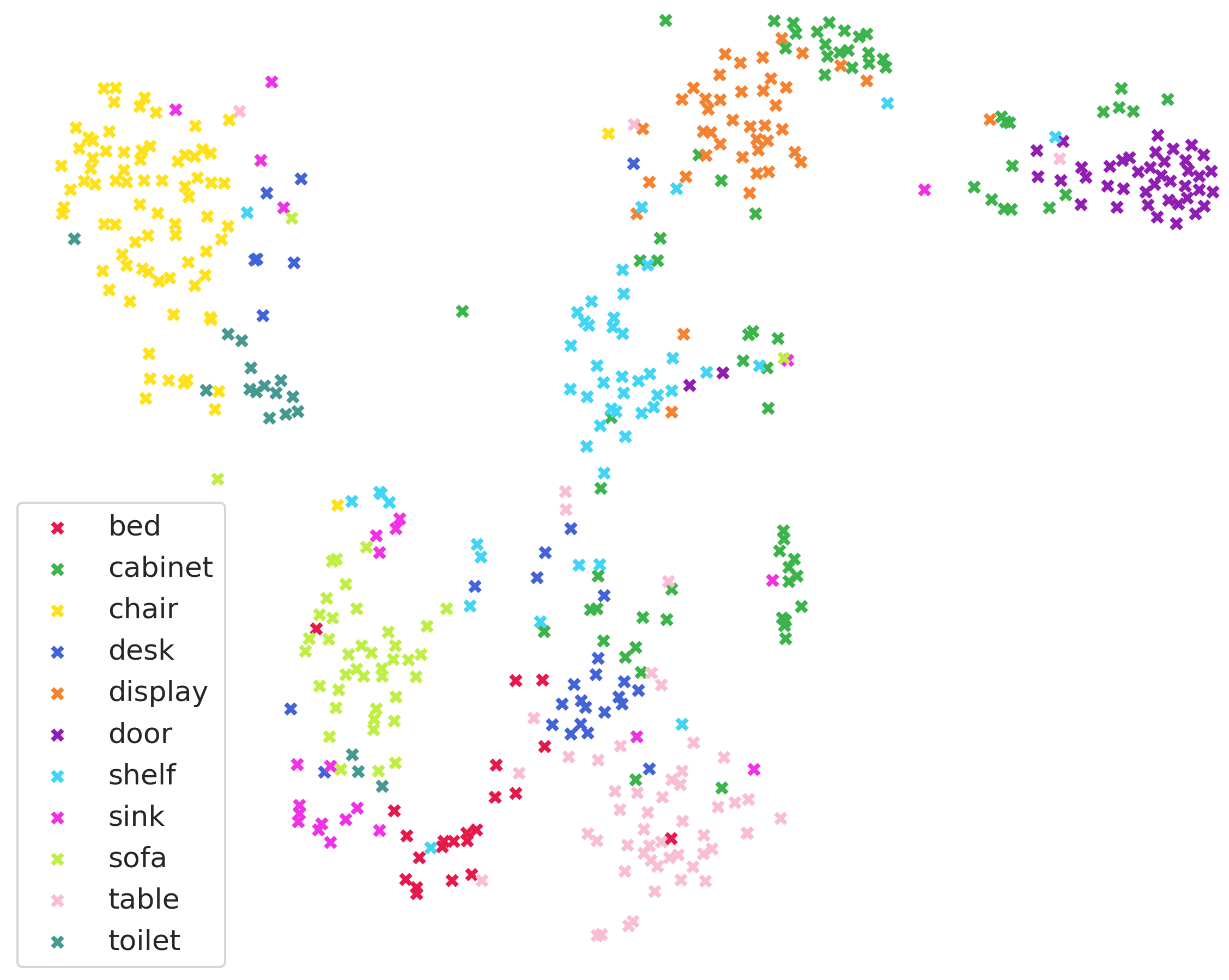}
    \end{minipage}}
    \subfigure[Ours: M11 $\rightarrow$ SO*11]{
        \label{fig:m2r4}
        \begin{minipage}[b]{0.23\textwidth}
            \includegraphics[width=\textwidth]{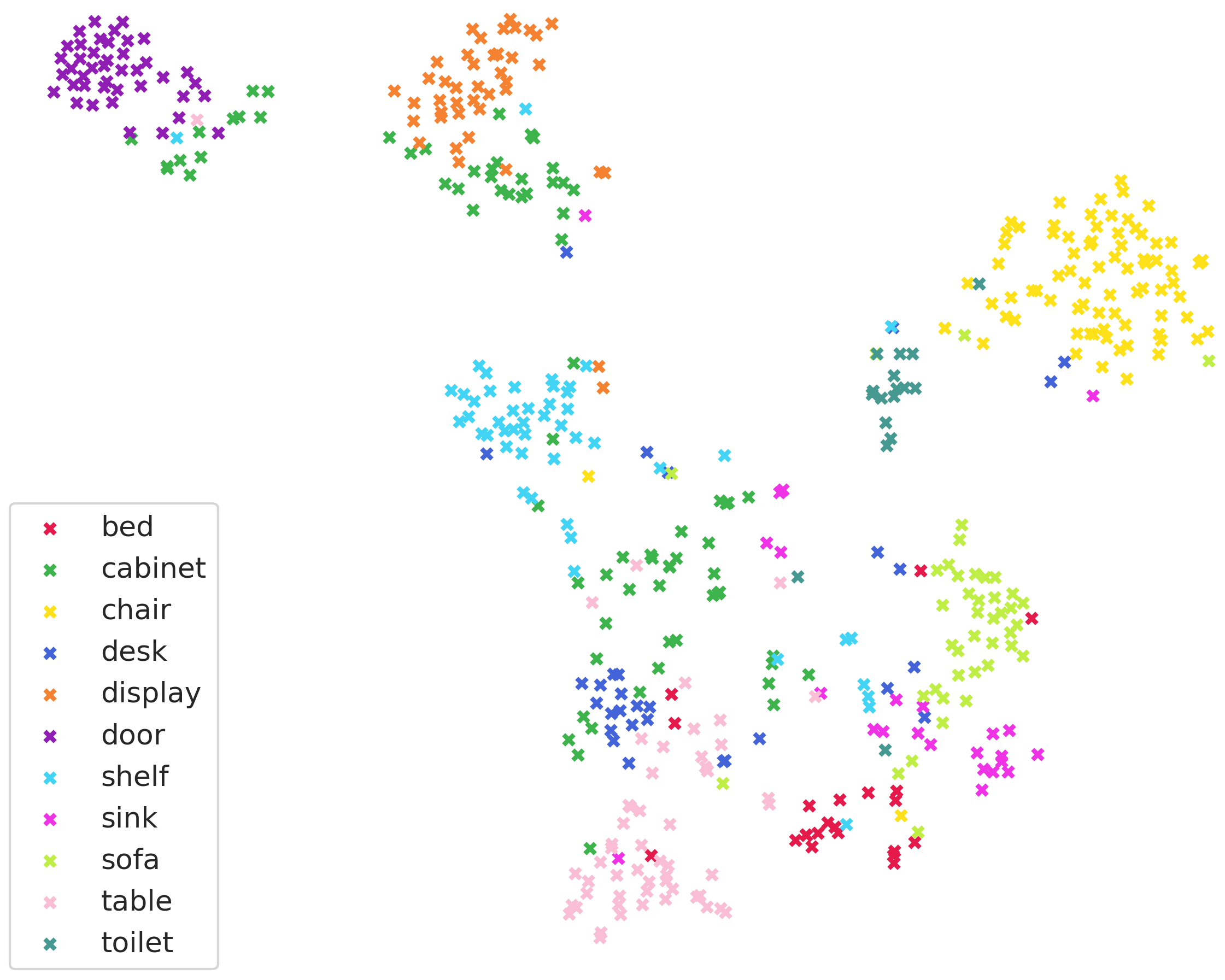}
        \end{minipage}}
    \subfigure[w/o Adapt: S10 $\rightarrow$ S*10]{
        \label{fig:s2r1}
        \begin{minipage}[b]{0.23\textwidth}
            \includegraphics[width=1\textwidth]{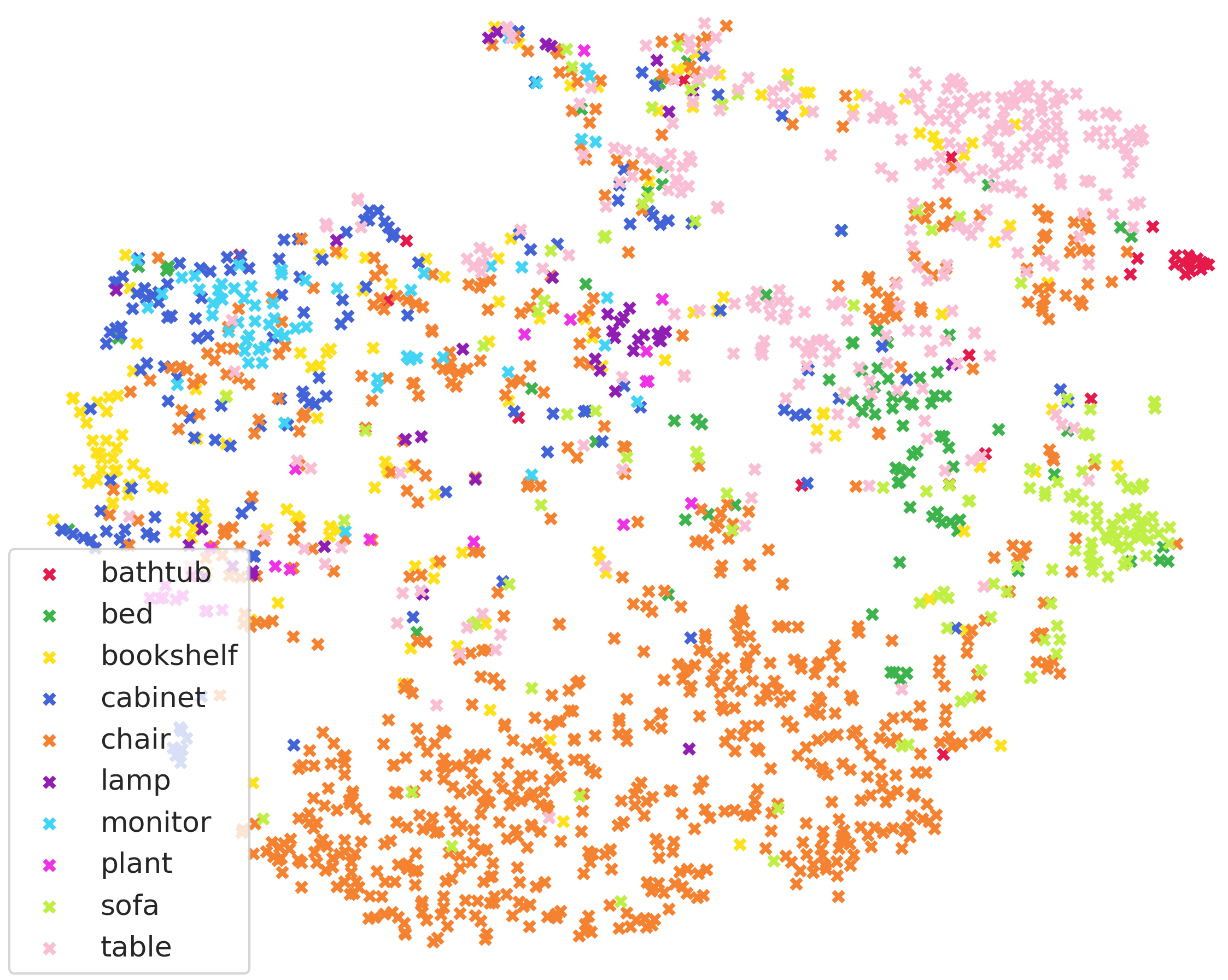}
        \end{minipage}}
    \subfigure[Ours: S10 $\rightarrow$ S*10]{
        \label{fig:s2r2}
        \begin{minipage}[b]{0.23\textwidth}
            \includegraphics[width=\textwidth]{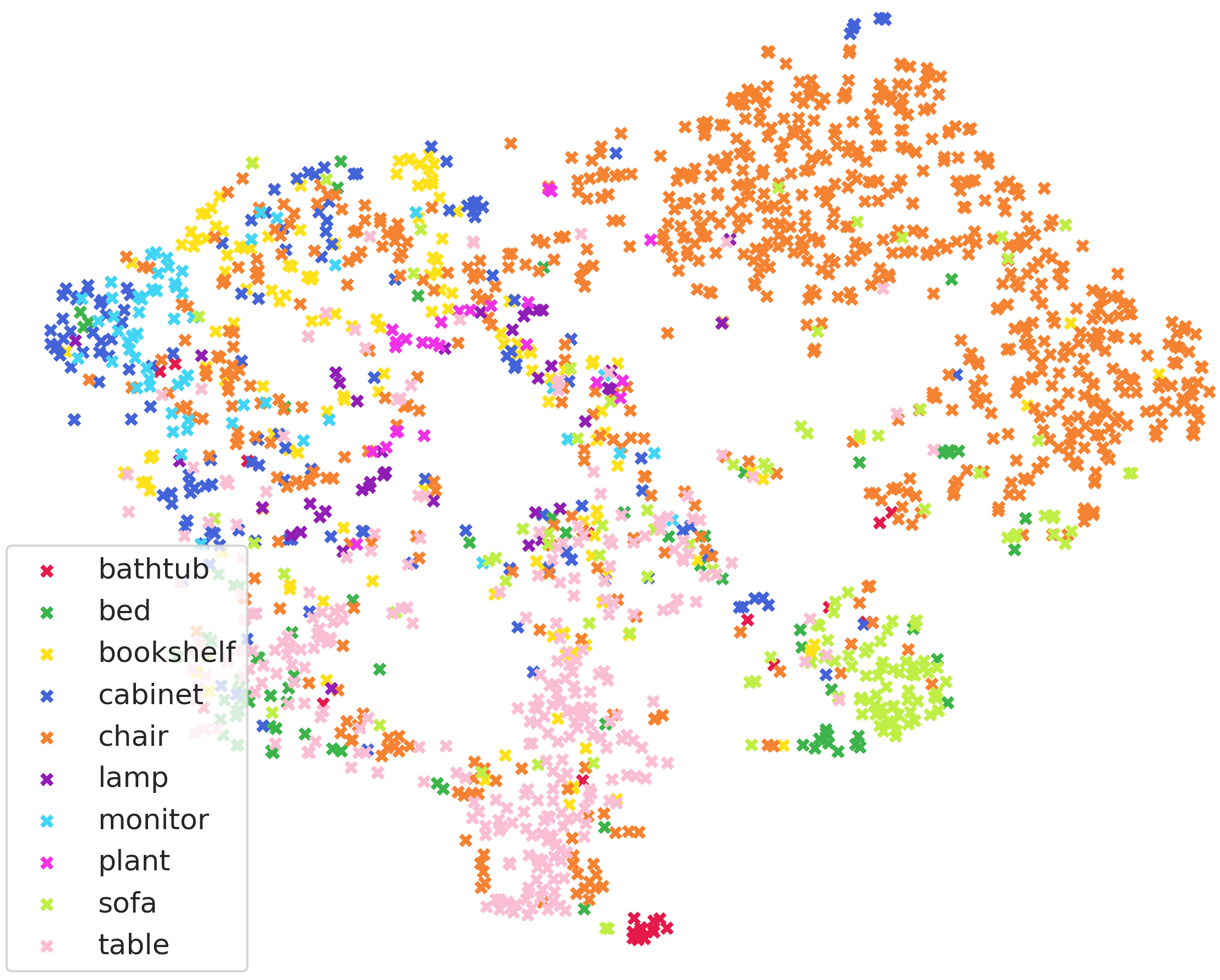}
        \end{minipage}}
    \subfigure[w/o Adapt: S9 $\rightarrow$ SO*9]{
        \label{fig:s2r3}
        \begin{minipage}[b]{0.23\textwidth}
            \includegraphics[width=1\textwidth]{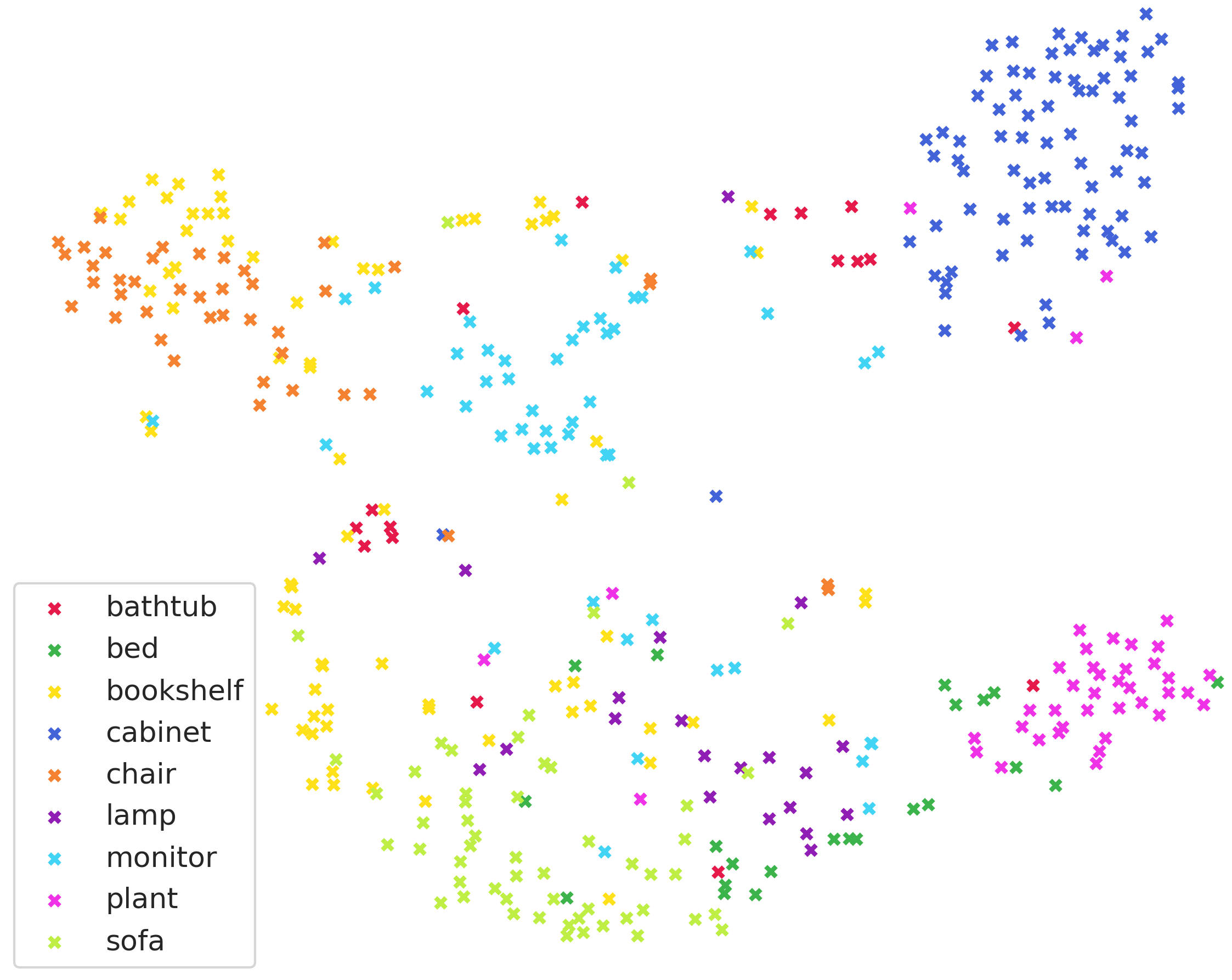}
        \end{minipage}}
    \subfigure[Ours: S9 $\rightarrow$ SO*9]{
        \label{fig:s2r4}
        \begin{minipage}[b]{0.23\textwidth}
            \includegraphics[width=\textwidth]{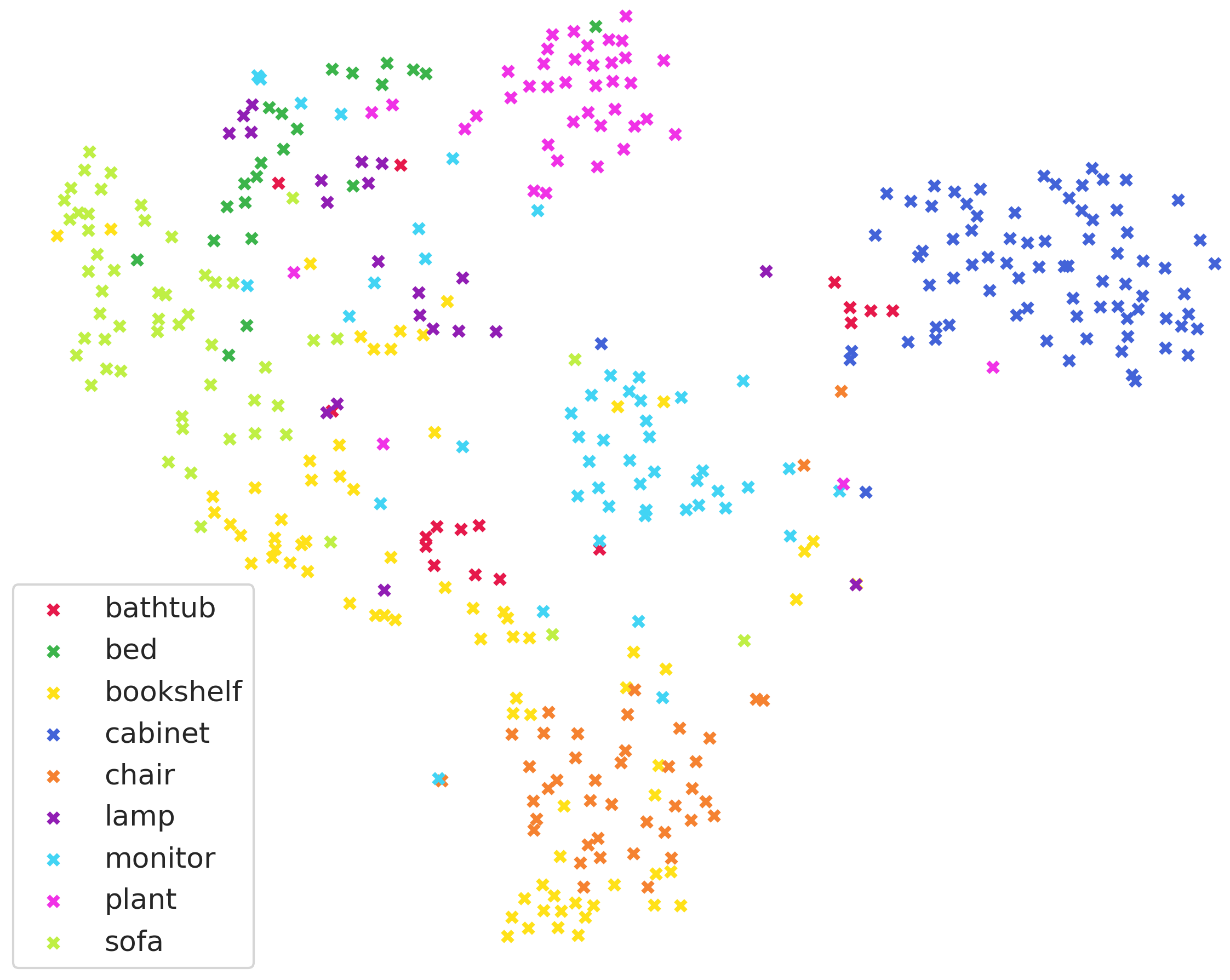}
        \end{minipage}}
    \caption{The t-SNE visualization of feature distribution on the target domain. Colors indicate different classes.}
    \label{Fig.embed_feature}
\end{figure*}

Visualization of confusion matrices in terms of class-wise classification accuracy achieved by {the w/o Adapt} and our TAM on the target domain of four Sim2Real UDA scenarios, \ie M10 $\rightarrow$ S*10, S10 $\rightarrow$ S*10, M11 $\rightarrow$ SO*11 and SO9 $\rightarrow$ SO*9, shown in Fig. \ref{Fig.confusion_matrix}. 
Furthermore, we utilize t-SNE \cite{t-SNE} to visualize the feature distribution on the target domain of four Sim2Real UDA scenarios. As shown in Fig. \ref{Fig.embed_feature}, the feature distribution of our proposed domain adaptation method on the target domain is more discriminative than that of the baseline without domain adaptation.

\vspace{-0.2cm}
\subsection{Discussion}
\noindent \textbf{Effectiveness and Adaptability:} The proposed TAM is particularly suited for Sim2Real scenarios due to the increased local variability in real-world data, such as noise and uneven sampling, compared to Sim2Sim settings. High-frequency signals effectively capture stable global geometry that remains consistent across both domains, while local implicit fields robustly handle unpredictable local variations. This dual capability enhances the model's adaptability and performance in transitioning from simulation to real environments, where local inconsistencies are more pronounced.

\noindent \textbf{Noise Robustness:} We evaluated our model's robustness under the ``Strong'' noise level, which includes jitter noise along with 30 outliers, and compared it with other methods. As shown in Table \ref{tab:noise_compare}, our approach demonstrates superior resistance to noise. Even with a high number of outliers and significant jitter noise, it maintains higher classification accuracy than competing methods, highlighting its effectiveness in handling noisy point cloud data.

\noindent \textbf{Computational Complexity:} We trained our model using an NVIDIA GeForce RTX 3090 Ti GPU. As shown in Table \ref{tab:param_anal}, we provide a clear comparison between our method and other baseline methods in terms of parameters, computation cost, and training time. Our method outperforms GAST \cite{GAST} and GAI \cite{GAI} with fewer parameters and lower computational cost. While DefRec \cite{PCM_RegRecT}, PointDAN \cite{PointDAN}, and SD \cite{SD} have certain advantages in efficiency, they fall short in overall performance.

\noindent \textbf{Future Work:} While this study focuses on Unsupervised Domain Adaptation (UDA) for point cloud classification using relatively simple datasets with around 10 rigid object classes, its applicability to more complex scenarios remains an open challenge. Specifically, the current setting, where 1024 points—sampled via farthest point sampling and normalized—are sufficient for representation, may not generalize well to large-scale and structurally diverse point clouds. In future work, we aim to extend our approach to UDA for large-scale 3D scene segmentation, addressing key challenges such as handling high-density point distributions, capturing fine-grained spatial structures, and improving scalability in terms of both computational efficiency and data complexity.

\begin{table}[H]
  \centering
  \caption{Comparison of classification accuracy among different methods under the "Strong" noise level, where the red-colored values indicate the lowest decrease.}
  \label{tab:noise_compare}
  \resizebox{\linewidth}{!}{
      \begin{tabular}{l||cccc}
      \toprule
      Method & M10$\rightarrow$S*10 & S10$\rightarrow$S*10 & M11$\rightarrow$SO*11 & S9$\rightarrow$SO*9\\
      \midrule
      w/o Adaptation & 25.2\textcolor{green}{$\downarrow$18.6} & 27.5\textcolor{green}{$\downarrow$15.0} & 40.83\textcolor{green}{$\downarrow$20.85} & 34.21\textcolor{green}{$\downarrow$23.21}\\
      \midrule
      PointDAN \cite{PointDAN} & 26.7\textcolor{green}{$\downarrow$18.1} & 30.3\textcolor{green}{$\downarrow$15.4} & 45.65\textcolor{green}{$\downarrow$17.67} & 38.41\textcolor{green}{$\downarrow$16.54}\\
      GLRV \cite{GAST} & 39.3\textcolor{green}{$\downarrow$21.1} & 42.8\textcolor{green}{$\downarrow$14.9} & 52.14\textcolor{green}{$\downarrow$25.91} & 44.52\textcolor{green}{$\downarrow$17.94}\\
      \midrule
      Ours  & \textbf{46.2}\textcolor{red}{$\downarrow$15.9} & \textbf{45.1}\textcolor{red}{$\downarrow$14.2} & \textbf{60.61}\textcolor{red}{$\downarrow$15.58} & \textbf{49.25}\textcolor{red}{$\downarrow$14.6}\\
      \bottomrule
      \end{tabular}
      }
\end{table}

\begin{table}[H]
\centering
\caption{Comparative analysis of training costs of different methods}
\begin{tabular}{lccc}
\hline
Method & Parameters(M) & FLOPs(G) & Training Time(H) \\ 
\hline
DefRec \cite{PCM_RegRecT} & 2.08 & 2.77 & 8.3 \\
PointDAN \cite{PointDAN} & 2.84 & 0.94 & 3.2 \\
SD \cite{SD} & 3.47 & 0.92 & 8.5 \\
GAI \cite{GAI} & 22.68 & 3.58 & 10.4 \\
GAST \cite{GAST} & 23.60 & 2.17 & 12.7 \\
TAM (ours) & 14.5 & 2.09 & 10.6 \\ 
\hline
\end{tabular}
\label{tab:param_anal}
\end{table}
\section{Conclusion}
This paper proposes a novel scheme for Sim2Real unsupervised domain adaptation on object point cloud classification that mitigates domain differences with topology-aware modeling. We demonstrate that global spatial topology and the topological relations within local description significantly enhance cross-domain generalization in Sim2Real scenarios.
In particular, our analysis reveals that the global spatial topology, unveiled through global high-frequency 3D spatial structures captured by Fourier Positional Encoding, showcases remarkable cross-domain robustness. The topological relations of local geometric implicits captured by our innovative self-supervised learning task also exhibit strong generalization capabilities.
Moreover, we introduce a novel self-training strategy combined with contrastive learning to mitigate the impact of noisy pseudo-labels, further enhancing the generalization ability of the representation.
Experiments on three Sim2Real benchmarks verified the effectiveness of our method, achieving the new state-of-the-art performance.


\bibliographystyle{IEEEtran}
\bibliography{IEEEabrv, sim2real.bib}

\vfill

\end{document}